\documentclass[sigconf,screen,nonacm,balance=false]{acmart}

\usepackage{booktabs} 

\setcitestyle{nosort,square} 

\usepackage{amsmath}
\DeclareMathOperator*{\argmax}{arg\,max} 

\usepackage[ruled]{algorithm2e} 

\SetAlFnt{\small}
\SetAlCapFnt{\small}
\SetAlCapNameFnt{\small}
\SetAlCapHSkip{0pt}
\usepackage{svg}
\usepackage{multirow}

\usepackage{dashrule}
\usepackage{xcolor}
\usepackage[outline]{contour}
\contourlength{0.6pt}
\contournumber{27}

\newcommand{\rev}[1]{{#1}}

\begin{document}
\title{NeRF as a Non-Distant Environment Emitter in Physics-based Inverse Rendering}

\author{Jingwang Ling}
\email{lingjw20@mails.tsinghua.edu.cn}
\affiliation{
 \institution{Tsinghua University}
 \country{China}
}

\author{Ruihan Yu}
\email{auroraryan0301@gmail.com}
\affiliation{
 \institution{Tsinghua University}
 \country{China}
}

\author{Feng Xu}
\email{xufeng2003@gmail.com}
\authornote{Corresponding author.}
\affiliation{
 \institution{Tsinghua University}
 \country{China}
}

\author{Chun Du}
\email{duchun@utibet.edu.cn}
\affiliation{
 \institution{Tibet University}
 \country{China}
}

\author{Shuang Zhao}
\email{shz@ics.uci.edu}
\affiliation{
 \institution{University of California, Irvine}
 \country{USA}
}

\begin{abstract}
    Physics-based inverse rendering enables joint optimization of shape, material, and lighting based on captured 2D images.
    To ensure accurate reconstruction, using a light model that closely resembles the captured environment is essential.
    Although the widely adopted distant environmental lighting model is adequate in many cases, we demonstrate that its inability to capture spatially varying illumination can lead to inaccurate reconstructions in many real-world inverse rendering scenarios.
    To address this limitation, we incorporate NeRF as a non-distant environment emitter into the inverse rendering pipeline. Additionally, we introduce an emitter importance sampling technique for NeRF to reduce the rendering variance.
    Through comparisons on both real and synthetic datasets, our results demonstrate that our NeRF-based emitter offers a more precise representation of scene lighting, thereby improving the accuracy of inverse rendering.
\end{abstract}

\begin{teaserfigure}
    \centering
        \begin{minipage}{0.03\textwidth}
            \centering
            \raisebox{\dimexpr0.5\baselineskip-\height}{\rotatebox{90}{\bf Hamster}}
        \end{minipage}%
        \begin{minipage}{0.97\textwidth}
            \includeinkscape[width=.98\linewidth]{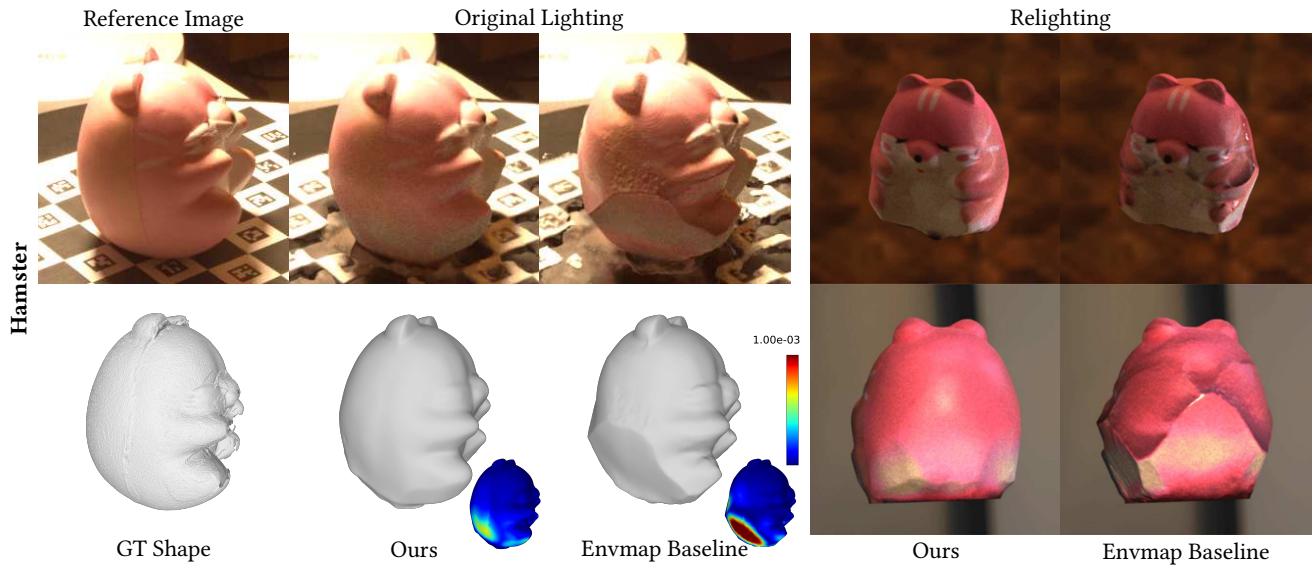}
        \end{minipage}
    \caption{We perform physics-based inverse rendering on real captured data featuring non-distant lighting. In this scenario, the commonly used environment map inaccurately models the lighting, leading to artifacts visible in relighting and shape reconstruction results. In contrast, our proposed NeRF emitter accurately models non-distant lighting, achieving high-quality inverse rendering.}
    \label{fig:teaser}
\end{teaserfigure}



\maketitle

\section{Introduction}
\label{sec:introduction}

Reconstructing the \rev{shape and material} of an object \rev{as well as the lighting condition} from 2D images has been a long-standing challenge in computer graphics and vision, offering numerous applications in \rev{3D reconstruction and scene digitization.}  %
Thanks to recent advancements, physics-based inverse rendering has gained popularity for its ability to %
accurately simulate light %
\rev{transport} within %
\rev{complex environments} %
\rev{and to produce high-quality reconstructions.}
\rev{%
    Also known as \emph{analysis by synthesis}, inverse rendering is typically formulated as optimization problems that seek for shape, material, and lighting parameters to minimize the difference (measured with some predetermined \emph{loss}) between captured input images and renderings generated with these parameters.
    To solve these optimizations efficiently using gradient-based methods (such as stochastic gradient descent), it is desired to incorporate \emph{differentiable rendering} in inverse-rendering pipelines.
}

To accurately simulate light transport for inverse rendering, having an accurate scene model is crucial.
However, in commonly used object-centric capture scenarios, the limitation in the number of captured images and the relatively localized camera positions pose challenges in reconstructing the entire scene.
As a result, inverse rendering approaches tend to model the object of interest in terms of shape and material, while approximating the rest of the scene as \rev{a distant ``environment''}.
The environment map \cite{DBLP:conf/siggraph/DebevecM97} has thus become a popular method for approximating lighting around objects. 
However, we demonstrate that in situations where the light source is not infinitely distant, an environment map proves to be an inadequate approximation, leading to inaccurate inverse-rendering results.

\begin{figure}[t]
    \centering
    \includeinkscape[width=\linewidth]{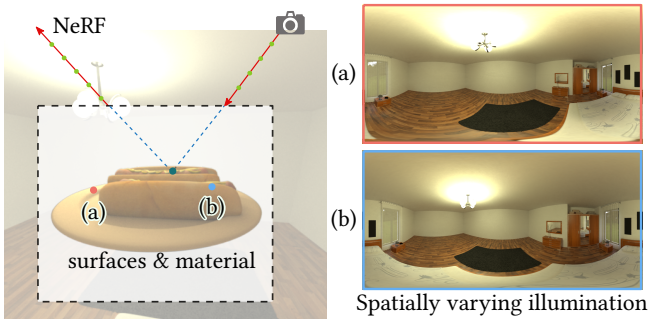}
    \caption{The region within the bounding box is modeled by surfaces and material, while the region outside is handled by NeRF to account for environmental lighting. The NeRF-synthesized illumination viewed from two shading points (red (a) and blue (b) dots) is visualized on the right.}
    \label{fig:pipeline}
\end{figure}

The crucial issue lies in the distant lighting assumption of the environment map, where the lighting distribution is considered spatially invariant.
However, when the light source is not distant, it exhibits strong parallax effects, as the light arrives from various directions at different positions on the object's surface (see Figure~\ref{fig:pipeline} right).
Consequently, a spatially invariant lighting model becomes an inadequate approximation for the actual lighting conditions.
To effectively model non-distant lighting, a representation capable of synthesizing spatially varying incoming radiance distribution is necessary.

We have %
\rev{observed} that a neural radiance field (NeRF) \cite{DBLP:conf/eccv/MildenhallSTBRN20} is well-suited for representing real, spatially varying lighting.
While an environment map essentially models a 2D radiance field at an infinite distance, NeRF models a radiance field that resides within 3D non-distant volumetric densities.
While NeRF is originally used for novel-view synthesis, we demonstrate its extension to light modeling, as moving the shading point is analogous to moving the viewpoint.
With an HDR NeRF representing the unbounded scene surrounding the object, we can synthesize 3D-consistent incoming radiance at any shading point on the object.

Therefore, to achieve more accurate inverse rendering when the lighting is not distant, we propose a technique that utilizes a NeRF to model the environmental lighting. %
We model the scene in a hybrid manner, with the object \rev{of interest} represented by \rev{a surface} %
\rev{and the surrounding by a NeRF}.
\rev{%
    To render these hybrid scenes, we generalize the standard surface-only rendering equation to use NeRF-based illumination.
    Additionally, we introduce an importance sampling scheme for NeRFs.
}
\rev{Then}, we build an inverse rendering pipeline that effectively reconstructs shape, material, and NeRF lighting from captured images.

We capture both real and synthetic datasets featuring non-distant lighting to compare our method with a %
\rev{baseline} pipeline that employs environment map emitters.
Our results demonstrate that an environment map inaccurately models non-distant lighting, resulting in artifacts in inverse-rendering results.
In contrast, our proposed NeRF emitter accurately models non-distant lighting and achieves high-quality %
\rev{reconstructions}.

\section{Related Work}

{\it Neural Radiance Fields (NeRF)} \cite{DBLP:conf/eccv/MildenhallSTBRN20} employ neural networks to model the radiance field in 3D scenes, enabling novel-view synthesis. 
Mip-NeRF 360 \cite{DBLP:conf/cvpr/BarronMVSH22} proposes a scene contraction method that extends NeRF to unbounded scenes, enabling NeRF to model the surround environment of an object of interest.
Instant-NGP \cite{DBLP:journals/tog/MullerESK22} accelerates NeRF by proposing a hybrid grid-network field representation.
NeRFStudio \cite{DBLP:conf/siggraph/TancikWNLYWKASA23} and Zip-NeRF \cite{DBLP:conf/iccv/BarronMVSH23} combine the works by \citet{DBLP:conf/cvpr/BarronMVSH22,DBLP:journals/tog/MullerESK22} to accelerate the modeling of an unbounded environment using NeRF.
RawNeRF \cite{DBLP:conf/cvpr/MildenhallHMSB22} and VR-NeRF \cite{DBLP:conf/siggrapha/XuALGB0BPKBLZR23} explore extending NeRF to high dynamic range, thereby improving the novel-view synthesis quality in dark-light or virtual reality scenarios. Considering that light sources in physics-based rendering are often HDR, integrating HDR technique into NeRF holds the potential to represent an HDR environment emitter.

Based on the powerful modeling capabilities of neural networks, several research efforts have delved into leveraging neural networks to represent complex luminaires.
The works by \citet{DBLP:journals/tog/ZhuBXBV0SHY21,DBLP:conf/rt/CondorJ22} employ neural networks to model the light field or radiance field of virtual luminaries, thereby accelerating physics-based rendering by reducing variance.
DMRF \cite{DBLP:conf/iccv/QiaoGXF0L23} explores the insertion of a virtual mesh into the trained NeRF and simulates the interactions between them.
While these works focus on using pretrained networks to model light in forward rendering, our work considers employing NeRF in inverse rendering as an emitter to reconstruct scene parameters.

{\it Differentiable Rendering} involves differentiating the rendering equation to obtain gradients of scene parameters, which is the foundation of gradient-based optimization in the subsequent inverse rendering.
A key challenge in differentiable rendering lies in the differentiation of visibility discontinuities, which is essential to enable shape optimization.
The works by \citet{DBLP:journals/tog/LiADL18,DBLP:journals/tog/ZhangWZGRZ19,DBLP:journals/tog/ZhangMYGZ20} explicitly sample edges at the silhouettes of shapes to estimate the boundary term that arises when differentiating the integral over visibility discontinuities.
Another approach, based on warped-area sampling \cite{DBLP:journals/tog/BangaruLD20}, sidesteps visibility discontinuities by converting boundary integrals into area integrals.
This approach can be extended from meshes to signed distance functions (SDF) to enable differentiable SDF rendering \cite{DBLP:journals/tog/ViciniSJ22,DBLP:conf/siggrapha/BangaruGLLSHBXB22}, and has also recently been extended to path space to differentiate path integrals \cite{DBLP:journals/tog/XuBLZ23}.
Deriving differentiable volume rendering has been previously explored by \citet{DBLP:journals/tog/ZhangWZGRZ19}, but our model distinguishes itself by incorporating emissive volume and excluding scattering.

{\it Inverse Rendering} aims to utilize gradient descent to minimize rendering loss between rendered and captured images to reconstruct the scene shape, material, and lighting parameters.
Accurate lighting models are crucial to achieve high-quality reconstruction via inverse rendering.
However, current inverse rendering methods often rely on simplistic light models, which either restrict the capture setup or lead to inaccurate approximations of real-world scene lighting.
Some works require specific capture configurations in controlled environments, such as a co-located camera-light setup \cite{DBLP:conf/eccv/BiXSHHKR20,DBLP:journals/corr/abs-2008-03824,DBLP:conf/cvpr/ZhangLLS22} or a point light source \cite{DBLP:conf/eccv/YangCCCW22,DBLP:conf/nips/YangCCCW22,DBLP:conf/cvpr/Ling0023}. 
For inverse rendering in ordinary environments, an environment map \cite{DBLP:conf/siggraph/DebevecM97} commonly serves as a lighting approximation.
Numerous works utilize inverse rendering for reconstructing individual objects \cite{DBLP:conf/cvpr/ZhangLWBS21,DBLP:conf/iccv/BossBJBLL21,DBLP:conf/nips/BossJBLBL21,DBLP:conf/nips/BossEKLSBLJ22}, outdoor buildings \cite{DBLP:journals/corr/abs-2112-05140}, and humans \cite{DBLP:conf/eccv/ChenL22}.
There is also a trend in inverse rendering to be more physically accurate by considering shadows \cite{DBLP:conf/nips/HasselgrenHM22,DBLP:conf/cvpr/SrinivasanDZTMB21}, interreflections \cite{DBLP:conf/cvpr/ZhangSHFJZ22,DBLP:conf/cvpr/WuHLZF023,DBLP:conf/cvpr/JinLXZHBZX023}, and differentiable path tracing \cite{DBLP:conf/iccv/0004CLYZM0Z023}.
All of the above methods employ an environment map as the scene lighting model.
However, an environment map assumes that the scene lighting is infinitely distant, which is a rare case in real-world capture setups.
We demonstrate that using an environment map can result in degraded reconstruction quality when this assumption is violated.
Therefore, we propose a NeRF emitter module designed to be a universal component applicable across inverse rendering systems, avoiding the distant lighting assumption.

\newcommand{\pixel}{I}
\newcommand{\image}{\widehat{I}}
\newcommand{\npixel}{N}
\newcommand{\Li}{L_i}
\newcommand{\Lo}{L_o}
\newcommand{\Le}{L_e}
\newcommand{\Lisurface}{\Li^s}
\newcommand{\Livolume}{\Li^v}
\newcommand{\Lisurfaceprime}{\Li'^s}
\newcommand{\Livolumeprime}{\Li'^v}
\newcommand{\campos}{\mathbf{x}_c}
\newcommand{\pos}{\mathbf{p}}
\newcommand{\pixelarea}{\mathcal{A}}
\newcommand{\importance}{W}
\newcommand{\surfacetwod}{\mathcal{S}^2}
\newcommand{\solidangle}{\mathbf{\omega}}
\newcommand{\solidangleprime}{\solidangle'}
\newcommand{\projectedsolidangle}{\solidangle^\perp}
\newcommand{\projectedsolidangleprime}{\solidangleprime^\perp}
\newcommand{\transport}{\mathbf{r}}
\newcommand{\transportdistance}{D}
\newcommand{\interdist}{t_0}
\newcommand{\dist}{t}
\newcommand{\transmittence}{T}
\newcommand{\density}{\sigma}
\newcommand{\bsdf}{f_s}
\newcommand{\radiance}{\mathbf{c}}
\newcommand{\maxcontribdist}{\dist_{\mathrm{max}}}
\newcommand{\luminance}{Y}
\newcommand{\nmixture}{M}
\newcommand{\mean}{\bar}
\newcommand{\gaussianmean}{\mathbf{\mu}}
\newcommand{\gaussiancovariancei}{\kappa_i^{-1}\mathbf{I}}
\newcommand{\gaussianweight}{\lambda}
\newcommand{\vmfmean}{\widehat{\mathbf{\mu}}}
\newcommand{\vmfkappa}{\widehat{\kappa}}
\newcommand{\vmfweight}{\widehat{\lambda}}
\newcommand{\lTwoNorm}[1]{\left\lVert #1 \right\rVert_2}
\newcommand{\lOneNorm}[1]{\left\lVert #1 \right\rVert_1}
\newcommand{\sceneparam}{\mathbf{x}}
\newcommand{\loss}{\mathcal{L}}
\newcommand{\lossrender}{\loss_{\mathrm{render}}}
\newcommand{\stopgrad}[1]{\mathrm{sg}(#1)}

\section{Preliminaries}

We revisit differentiable surface rendering \cite{DBLP:journals/tog/Nimier-DavidSRJ20} and NeRF \cite{DBLP:conf/eccv/MildenhallSTBRN20}, comparing their similarities and distinctions.
Image pixels $\pixel_1$, $\pixel_2$, ..., $\pixel_k$, ..., $\pixel_{\npixel}$ measure the integral of the dot product between the sensor importance function $\importance$ and the incident radiance $\Li$ at a position $\pos$:
\begin{equation}
    \pixel_k = \int_{\pixelarea} \int_{\surfacetwod} \importance_k(\pos, \solidangle) \Li(\pos, \solidangle) d\projectedsolidangle d\pos,
\end{equation}
where $\solidangle$ denotes the light direction and $\projectedsolidangle$ is the projected solid angle.
Surface and volume rendering exhibit differences in assumptions about the scene, affecting $\Li(\pos, \solidangle)$. 
Surface rendering assumes a vacuum between surfaces, where incoming radiance equals outgoing radiance at the first intersection point of the ray, thus $\Li$ is given by the surface version $\Lisurface$:
\begin{equation}
    \label{eq:Lisurfacepre}
    \Lisurface(\pos, \solidangle) = \Lo(\transport(\pos, \solidangle, \interdist), -\solidangle),
\end{equation}
where $\interdist$ is the distance to the closest intersection point $\transport(\pos, \solidangle, \interdist)$ along direction $\solidangle$.
Outgoing radiance $\Lo$ involves an integral over the hemisphere, expressed by the rendering equation:
\begin{equation}
    \label{eq:Lo}
    \Lo(\pos, \solidangle) = \Le(\pos, \solidangle) + \int_{\surfacetwod}\Li(\pos, \solidangleprime)\bsdf(\pos,\solidangle,\solidangleprime)d\projectedsolidangleprime,
\end{equation}
where $\Le$ represents the surface emitted radiance and $\bsdf$ denotes the BSDF.
The $\Li$ on the right-hand side involves recursive computation and can be implemented using path tracing.

On the other hand, NeRF makes different assumptions about the scene. 
It assumes that the scene is filled with emissive volume, without surfaces or volumetric scattering.
In this case, $\Li$ is determined by the volumetric version $\Livolume$, calculated using the (non-scattering) volume rendering equation:
\begin{equation}
    \label{eq:volume_rendering_pre}
    \Livolume(\pos, \solidangle) = \int_{0}^{\infty}\transmittence(\pos, \solidangle, \dist)\density(\transport(\pos,\solidangle,\dist))\radiance(\transport(\pos,\solidangle,\dist), -\solidangle)d\dist,
\end{equation}
where $\density$ represents the volumetric density, $\radiance$ is the emission, and $\transmittence$ denotes the transmittance, which models the absorption effect of occluding densities:
\begin{equation}
    \label{eq:transmittence}
    \transmittence(\pos, \solidangle, \dist) = \exp\left(-\int_{0}^{\dist}\density(\transport(\pos, \solidangle, s))ds\right).
\end{equation}

In differentiable surface rendering, we need to differentiate Equation~\ref{eq:Lo}.
Assuming we eliminate the discontinuities using techniques like reparameterization \cite{DBLP:journals/tog/BangaruLD20,DBLP:journals/tog/ViciniSJ22}, we obtain
\begin{align}
    \label{eq:diff_Lo}
\begin{split}
    \partial_\sceneparam\Lo(\pos, \solidangle) = \partial_\sceneparam\Le(\pos, \solidangle) +\int_{\surfacetwod} [&\partial_\sceneparam\Li(\pos, \solidangleprime)\bsdf(\pos,\solidangle,\solidangleprime) \\
    +&\Li(\pos, \solidangleprime)\partial_\sceneparam\bsdf(\pos,\solidangle,\solidangleprime)]d\projectedsolidangleprime,
\end{split}
\end{align}
where $\partial_\sceneparam\Li(\pos, \solidangleprime)$ represents the derivative provided by the light source when it is directly visible from the direction $\solidangleprime$.

The differentiation of NeRF is straightforward.
Equation~\ref{eq:volume_rendering_pre} can be computed via ray marching and is trivially differentiable. 
Gradients can be computed using automatic differentiation such as PyTorch \cite{DBLP:conf/nips/PaszkeGMLBCKLGA19}.

\section{NeRF-based Non-distant Emitters}

\subsection{Hybrid Rendering of Surfaces and NeRF}
\label{sec:primal_rendering}

We propose a NeRF emitter to model the environment lighting surrounding the object.
We adopt a hybrid scene representation containing surfaces and NeRF.
As depicted in Figure~\ref{fig:pipeline} left, we partition the scene into two regions based on the bounding box of the object: the internal region is represented by surface and material, while the external region is represented by the NeRF-based environmental lighting.
Rendering the hybrid scene involves a rendering equation that accounts for both surfaces and NeRF.
According to our scene assumption, the incoming radiance $\Li$ consists of $\Lisurfaceprime$ from surfaces and $\Livolumeprime$ from the NeRF:
\begin{equation}
    \label{eq:Li}
    \Li(\pos, \solidangle) = \Lisurfaceprime(\pos, \solidangle) + \Livolumeprime(\pos, \solidangle).
\end{equation}
Considering the surface contribution $\Lisurfaceprime$, it follows the surface light transport, with additional consideration of the occluding NeRF:
\begin{equation}
    \label{eq:Lisurface}
    \Lisurfaceprime(\pos, \solidangle) = \transmittence(\pos, \solidangle, \interdist)\Lo(\transport(\pos, \solidangle, \interdist), -\solidangle),
\end{equation}
where $\transmittence$ represents the NeRF transmittance defined in Equation~\ref{eq:transmittence}.
The NeRF contribution $\Livolumeprime$ resembles Equation~\ref{eq:volume_rendering_pre}, but with zero densities within the bounding box. 
Furthermore, we integrate from 0 to $\interdist$ instead of $\infty$, where $\interdist$ denotes the distance to the first surface intersection. $\interdist$ goes to $\infty$ when there is no surface intersection.
The $\Li$ defined in Equation~\ref{eq:Li} is used recursively in the computation of Equation~\ref{eq:Lo} in path tracing.
When the light path eventually exits the bounding box and goes to infinity, NeRF is queried to calculate the incoming radiance from environment lighting, akin to the behavior of a conventional light source.

\rev{
Physics-based rendering necessitates its components, such as NeRF in our case, to adhere to physical correctness.
While not often used in a physics-based context, we recognize that NeRF, fundamentally representing a radiance field, is well suited for modeling the physical quantity of ``radiance'' within a scene. 
NeRF simplifies light transport outside the bounding box by considering interreflections as emissions, yet the light paths within the bounding box are preserved.
Although this simplification cannot capture the illumination changes resulting from scene edits after reconstruction, it effectively models interreflections during the capture phase, hence suitable for inverse rendering.
}

\rev{
Our formulation allows for arbitrary bounding box delineation of the regions. 
We allow users to place a bounding box to denote the object of their interest.
Light sources within the bounding box are accounted for with the self-emission $\Le$ in Equation~\ref{eq:Lo}. 
Any part of the scene not enclosed within the bounding box is treated as part of the environmental light and modeled using NeRF. 
In the experiments, to mitigate inverse rendering ambiguity, we introduce a prior by constraining emission within the bounding box to zero. 
Imposing this prior however requires all emissive objects, such as area lights, to be located outside the bounding box. 
This prior is optional and not a fundamental limitation.
}

The scene properties also simplify NeRF computations to at most two rays in a light path.
Assuming consistently zero densities within the bounding box, and given its convex nature, any rays originating and terminating within it remain inside, with zero densities along them.
This leaves only the necessity to compute NeRF for the two rays entering or exiting the bounding box.

\subsection{Emitter Importance Sampling for NeRF}
\label{sec:monte_carlo}
    
\begin{figure}[t]
    \centering
    \includeinkscape[width=.98\linewidth]{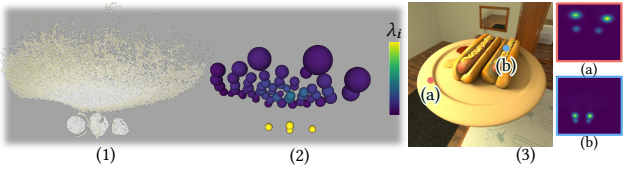}
    \caption{We generate importance sampling distributions by (1) creating a point cloud of the bright parts of the NeRF, (2) clustering it into Gaussian mixtures, and (3) projecting these Gaussians to vMFs at shading points (red (a) and blue (b) dots).}
    \label{fig:cluster}
\end{figure}

To integrate the NeRF emitter into physically-based rendering, Monte Carlo sampling is required. 
This entails the ability to sample directions and compute the corresponding probability densities.
Importance sampling for NeRF presents unique challenges due to its nature as a volumetric light source governed by a density field with without fixed topology.
Hence, a specific importance sampling strategy is imperative for NeRF.

Our basic idea is to approximate regions of significant radiance contribution in NeRF using simple geometric primitives.
Specifically, we employ a Gaussian mixture model to fit a point cloud extracted from the bright regions of NeRF, acting as a proxy model during importance sampling.
First, we randomly sample rays originating from positions inside the bounding box. We use Equation~\ref{eq:volume_rendering_pre} to compute the radiance along this ray and record the depth value that maximally contributes to the rendered radiance
\begin{equation}
    \maxcontribdist=\argmax_{\dist}\transmittence(\pos, \solidangle, \dist)\density(\transport(\pos,\solidangle,\dist))\radiance(\transport(\pos,\solidangle,\dist), -\solidangle).
\end{equation}
Then we add a sample point to the point cloud at the position $\transport(\pos, \solidangle, \maxcontribdist)$ with the weight $\luminance(\Livolume(\pos,\solidangle))$, where $\luminance$ is the operator that converts RGB radiance to monochrome luminance.
We aim to approximate the radiance distribution of the point cloud via fitting an isotropic Gaussian mixture model of $\nmixture=64$ components. 
To efficiently minimize rendering variance using a limited number of components, inspired by MIS compensation \cite{DBLP:journals/tog/KarlikSVSK19}, we subtract the mean weight $\mean{\luminance}(\Livolume(\pos,\solidangle))$ from the weights of all the points. Only points with positive weights are retained.
As a result, the remaining points (Figure~\ref{fig:cluster} (a)) can focus on accurately capturing the shape of very bright regions, while remaining unbiased via the multiple-importance sampling combined with BSDF sampling.
Based on the compensated weights, we cluster the point cloud into $\nmixture$ isotropic Gaussians  (Figure~\ref{fig:cluster} (b)), each characterized by a mean $\gaussianmean_i$, covariance $\gaussiancovariancei$ and weight $\gaussianweight_i$, with $i\in [1, \nmixture]$.
When performing emitter importance sampling for a shading point $\pos$, we first project the isotropic Gaussians into von Mises–Fisher distributions $(\vmfmean_i, \vmfkappa_i, \vmfweight_i)$:
\begin{equation}
    \vmfmean_i = \frac{\gaussianmean_i-\pos}{\lTwoNorm{\gaussianmean_i-\pos}},\; \vmfkappa_i=\frac{\kappa_i}{\lTwoNorm{\gaussianmean_i-\pos}}, \; \vmfweight_i=\gaussianweight_i,
\end{equation}
and use the projected vMF lobes  (Figure~\ref{fig:cluster} (c)) to perform sampling according to the work by \citet{jakob2012numerically}.
In inverse rendering, as NeRF parameters update during joint optimization, the brightness distribution may change. To ensure the sampling distribution matches the updated NeRF, we periodically repeat the above process to establish an updated Gaussian mixture model.

\subsection{Differentiable Hybrid Rendering}
\label{sec:differentiable_rendering}

Differentiable rendering of a hybrid scene, as defined in Section~\ref{sec:primal_rendering}, requires specific considerations.
First, we must address $\partial_\sceneparam\Li(\pos, \solidangleprime)$ in Equation~\ref{eq:diff_Lo}, which calculates the gradient of the light with respect to an arbitrary scene parameter $\sceneparam$. 
As Equation~\ref{eq:diff_Lo} relies on detached sampling \cite{DBLP:journals/tog/ZeltnerSGJ21}, $\solidangleprime$ does not receive gradients and $\partial_{\solidangleprime}\Li(\pos, \solidangleprime)$ is therefore zero.
However, it remains crucial to compute $\partial_\pos\Li(\pos, \solidangleprime)$, essential for differentiating shape parameters.
Note that $\pos$ also serves as the ray starting position input to NeRF.
Therefore, we need to compute gradients of the NeRF radiance with respect to the ray starting position via automatic differentiation and propagate it to Equation~\ref{eq:diff_Lo}.

Second, when we differentiate the volume rendering equation, we obtain a boundary term due to the dependence of the upper integration limit, $\interdist$, on the scene's shape parameters:
\begin{align}
\begin{split}
    \partial_\sceneparam\Livolumeprime(\pos, \solidangle) = &\transmittence(\pos, \solidangle, \interdist)\density(\transport(\pos,\solidangle,\interdist))\radiance(\transport(\pos,\solidangle,\interdist), -\solidangle)\partial_\sceneparam \interdist \\
    + \int_{0}^{\interdist}\partial_\sceneparam(&\transmittence(\pos, \solidangle, \dist)\density(\transport(\pos,\solidangle,\dist))\radiance(\transport(\pos,\solidangle,\dist), -\solidangle))d\dist.
\end{split}
\end{align}
However, densities $\density$ approach zero continuously at the bounding box's boundaries and are zero within it. Since $\transport(\pos,\solidangle,\interdist)$ always stays within the bounding box, this boundary term always evaluates to zero.

\section{Inverse Rendering using NeRF Emitter}
Building upon the derivations above, we've implemented a pipeline that integrates NeRF as an emitter into physics-based inverse rendering. 
While previous inverse rendering methods typically require object foreground masks to exclude background influences, our pipeline can simultaneously obtain scene information from both foreground and background pixels and jointly optimize shape, material, and NeRF emitter based on the gradients of the rendering loss.
Physics-based inverse rendering often requires proper initialization. 
Therefore, we designed a multi-stage optimization process that utilizes NeRF to aid the initialization of inverse rendering and facilitate the optimization process.
Additionally, due to the widespread use of environment maps, existing datasets tend to avoid using nearby light sources, leading to a lack of thorough testing for scenes with close light sources.
We established a capture system to thoroughly test such scenarios.

\subsection{Multi-Stage Optimization}
Compared to surface-based inverse rendering, NeRF demonstrates robustness in parameter initialization.
Therefore, we first conduct ordinary NeRF training to initialize the scene, with the interior of the bounding box represented using either NeRF densities or NeuS \cite{DBLP:conf/nips/WangLLTKW21,DBLP:conf/siggraph/YarivHRVSSBM23,DBLP:conf/iccv/GeHZ0C23}. 
We then extract the geometry within the bounding box using truncated signed distance function (TSDF) fusion \cite{DBLP:conf/siggraph/CurlessL96} to initialize the object shape, and clear the densities inside the bounding box.
The object shape will be continually refined by inverse rendering.
At this point, NeRF has a reasonable initial lighting approximation, and the object's geometry has been initialized through NeRF.
Finally, we perform joint optimization via inverse rendering until convergence.

\subsection{Non-Distant Emitter Capture System}
To thoroughly test inverse rendering on scenes with non-distant emitters, we establish a capture setup to capture a dataset featuring such scenarios.
We additionally provide multiple lighting conditions by rotating the object to mitigate ambiguity under single-light conditions.
Each scene includes a nearby light source placed around the object.
We employ a turntable to rotate the object, capturing multi-view images with a camera for each rotation pose. 
While the camera's viewpoint remains focused on the object, lighting cues can be observed from background pixels, providing accurate lighting information. 
To accurately capture the lighting, we utilize a DSLR camera to capture HDR images.

\section{Implementation Details}

\subsection{HDR NeRF Parameterization and Training}
Our NeRF model builds upon the nerfacto proposed by \citet{DBLP:conf/siggraph/TancikWNLYWKASA23}.
To faithfully represent the emitter, it's essential to train NeRF to output HDR radiance values.
Our inverse rendering pipeline takes HDR images as input.
Following the approach in RawNeRF \cite{DBLP:conf/cvpr/MildenhallHMSB22}, we use an exponential activation function $\exp(x-5)$ as the output activation function for NeRF radiance.
We utilize the relative L1 Loss as the rendering loss function:
\begin{equation}
    \lossrender=\frac{1}{\npixel}\sum_{k=1}^{\npixel}\lOneNorm{\frac{\pixel_k-\image_k}{\stopgrad{\pixel_k}+\epsilon}},
\end{equation}
where $\pixel_k$ represents the rendered pixel, $\stopgrad{\cdot}$ indicates a stop-gradient operator, $\image_k$ is the pixel on the captured image, and $\epsilon=10^{-3}$ is a small number introduced to downweight excessively dark pixels.
We use gradient scaling \cite{DBLP:journals/corr/abs-2305-02756} to prevent NeRF from generating floaters.
We adopt the $L_\infty$ contraction proposed by \citet{DBLP:conf/siggraph/TancikWNLYWKASA23} to handle unbounded environmental lighting.

\subsection{Integrating NeRF into Differentiable Rendering}

Our inverse rendering pipeline is implemented by combining NeRFStudio \cite{DBLP:conf/siggraph/TancikWNLYWKASA23} and Mitsuba 3 \cite{jakob2022mitsuba3}.
While NeRFStudio utilizes neural networks to represent a NeRF, Mitsuba 3 is a megakernel-based differentiable renderer that cannot incorporate neural networks within a megakernel. Therefore, integrating these two systems necessitates special techniques.

We encapsulate NeRF and its importance sampling approach into Mitsuba 3 as an emitter module.
Importance sampling is executed within Mitsuba 3 using Dr.Jit \cite{DBLP:journals/tog/JakobSRV22}.
During light source evaluation, the ray starting positions and directions are transferred to PyTorch for NeRF computation. 
The computed radiance or derivatives are then passed back to Mitsuba 3.
Since only the emitter involves neural networks, we employ one-sample MIS \cite{DBLP:phd/us/Veach97} to allow emitter queries for a batch of paths to take place simultaneously.
We split the rendering operation into two megakernels, with the neural network operations occurring between them.
Specifically, the first kernel generates NeRF query rays, and the second kernel collects NeRF evaluations for pixel reconstruction.

Due to NeRF's large memory footprint, it supports only a limited ray batch size. 
However, typical differentiable rendering systems operate on an image, exceeding NeRF's maximum batch size capacity.
Therefore, we divide the rays into batches to be sequentially processed by NeRF.
To save memory, we do not store computation graphs, but instead recompute them during gradient computation.
During forward rendering, we conduct detached gradient NeRF evaluation, and record random seeds to enable the replay \cite{DBLP:journals/tog/ViciniSJ21} of random operations.
During back-propagation, we recover the computation graph for each batch of rays for automatic differentiation.

As the NeRF evaluation constitutes the runtime bottleneck, we have developed a multi-GPU training system to enhance its parallelism. The primary GPU hosts a Mitsuba 3 scene and performs surface rendering. For light evaluation and gradient computation, this GPU is tasked with distributing NeRF queries to the other GPUs, as well as collating computation results.

\subsection{Shape and Material Optimization}
Our differentiable surface rendering builds upon the differentiable SDF proposed by \citet{DBLP:journals/tog/ViciniSJ22}.
This implementation effectively handles visibility discontinuities, facilitating shape optimization. 
We represent the shape and material using voxel grids with a resolution of $256^3$, and conduct inverse rendering for 320 iterations.
In each iteration, we randomly sample 6 images from the dataset for training.
Following a coarse-to-fine optimization approach, we start optimization using $128^2$ resolution images and $64^3$ voxel grids, progressively scaling up images to $512^2$ and voxel grids to $256^3$ at iterations 128 and 256.
We use a Laplacian loss \cite{DBLP:journals/tog/ViciniSJ22} and a curvature loss \cite{DBLP:conf/cvpr/RosuB23} to maintain the smoothness of voxel grids and redistance the SDF after each iteration. 
Both shape and material voxel grids, along with the NeRF network, are optimized using the Adam optimizer \cite{DBLP:journals/corr/KingmaB14}, with a learning rate of 3e-3 for shape, 2e-2 for material and 1e-2 for NeRF. We employ 512 primal and 128 adjoint samples per pixel.

\subsection{Data Capture Details}

Our dataset features two distinctive characteristics: high dynamic range and multiple rotation poses.
To obtain HDR captures, we use a Canon 5D Mark III camera with bracketed exposures. 
Each HDR image is synthesized from seven differently exposed photographs using HDRutils \cite{DBLP:conf/eccv/HanjiZM20,DBLP:journals/tci/HanjiM23}. 
This synthesis process ensures that HDR images accurately record the physical radiance. 
Employing a turntable, we capture four rotational poses for each object, rotating it by 90 degrees each time.
Multi-view images were captured for each rotation, and their camera intrinsic and extrinsic parameters were determined using Metashape.
To align the four rotational poses, we placed two ChArUco boards during the capture: one on the turntable and the other beside it. These boards aided in estimating the transformation of the bounding box with respect to the scene.
Inspired by recent works that jointly optimize NeRF and camera transformations \cite{DBLP:conf/iccv/LinM0L21}, we incorporate joint optimization of rotation transformations during the NeRF pretraining stage to refine the estimated transformations.
In NeRF, we assigned a unique appearance embedding to each rotation pose to facilitate modeling of environment lighting variations as the object rotates.

\section{Experiments}
The datasets used are first introduced, followed by the comparison with the environment map emitter in the task of inverse rendering and the evaluation of our emitter importance sampling.
\begin{figure*}[t]
    \centering
    \begin{minipage}[t]{.48\textwidth}
        \centering
        \includeinkscape[width=.98\linewidth]{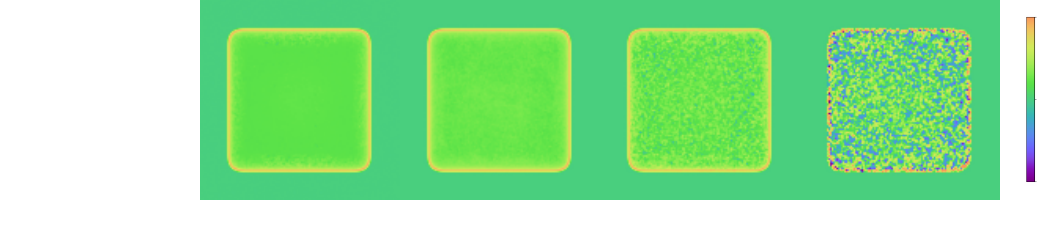}
        \includeinkscape[width=.98\linewidth]{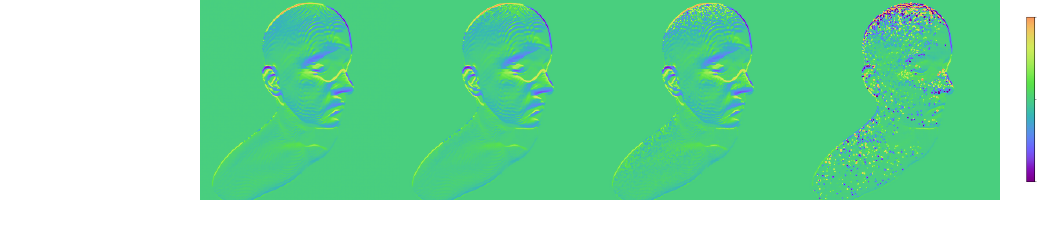}
    \end{minipage}
    \begin{minipage}[t]{.48\textwidth}
        \centering
        \includeinkscape[width=.98\linewidth]{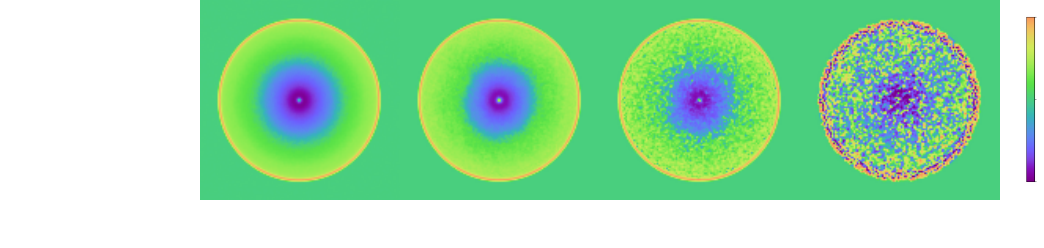}
        \includeinkscape[width=.98\linewidth]{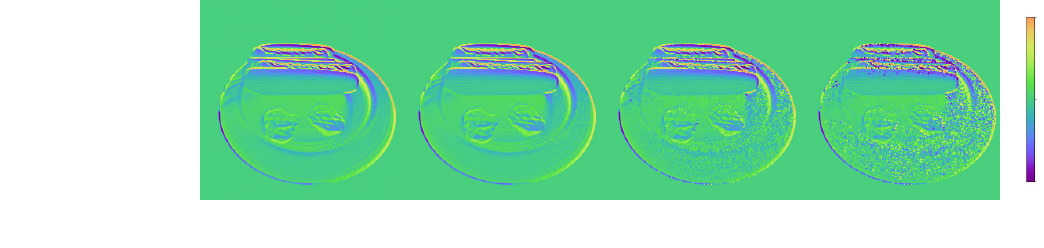}
    \end{minipage}
    \caption{Comparing the gradient images rendered by our emitter importance sampling and pure BSDF sampling.}
    \label{fig:gradient_image}
\end{figure*}\

\subsection{Datasets}
Following the aforementioned capture setup, we acquire four real-world data samples in an indoor environment.
We place a lamp beside the object to simulate non-distant lighting conditions.
Each sample consists of four rotation poses, with each rotation pose comprising 50-80 HDR multi-view images.
We use a Konica Minolta Vivid 9i scanner to obtain the ground truth geometry.
Following the real-world data capture setup, we synthesize four synthetic data samples using Mitsuba 3.
For each set of synthetic data, we place an object in an indoor scene and render it from four different rotation poses. For each rotation pose, we synthesize 50 multi-view images.

\subsection{Comparisons with Environment Map}

We compare the proposed NeRF emitter with the commonly employed environment map emitter, prevalent in recent inverse rendering frameworks  \cite{DBLP:journals/tog/ViciniSJ22,DBLP:conf/nips/BossEKLSBLJ22,DBLP:conf/cvpr/JinLXZHBZX023,DBLP:conf/iccv/0004CLYZM0Z023}.
Since the NeRF emitter is designed to be a universal component applicable across inverse rendering systems, in principle it could be integrated into these frameworks to achieve similar comparison results.
For fair assessment, we choose the differentiable SDF \cite{DBLP:journals/tog/ViciniSJ22} framework for its physics-based principles, and build both our method and the baseline upon it, thereby highlighting the differences stemming from the emitter.

We adapt the environment map baseline method to suit our data.
Firstly, we need an environment map input for the baseline. For synthetic data, we render environment maps using the ground truth scene in Mitsuba 3.
For real data, we train a NeRF with captured HDR images and render environment maps from the resulting NeRF.
In both cases, we place a virtual spherical camera at the center of the object and render environment maps.
When rendering the environment map, we retain the object of interest in the scene, and set the spherical camera's near plane at the object's bounding box. This approach captures global illumination effects caused by the object while preventing occlusion of the spherical camera.
We render an environment map for each rotation pose to capture the variations in global illumination caused by object rotations.

Secondly, the rendered background does not match the input images when using environment maps due to parallax effects. 
Therefore, for the synthetic dataset, we generate ground truth object masks and set the background pixels in the input images to black.
During inverse rendering, the baseline renders only the object and hides the environment map, thus generating masked rendered images that match the background-processed input images.
For the real dataset, we lend the background modeling ability of our NeRF emitter to the baseline.
Specifically, we use NeRF to render a background image behind the bounding box and a semi-transparent image to model occlusions (if any) in front of the bounding box.
These images are composited with the rendered object image to obtain the final image in the baseline.

Thirdly, the baseline employs a similar coarse-to-fine optimization procedure to ours. We use the same NeRF-extracted geometry as the shape initialization for the baseline.
After these adaptations, the key difference between our method and the baseline remains in different emitter models.

\begin{table}[t]\footnotesize
    \centering
    \caption{Quantitative comparison with the environment map baseline.}
    \label{tab:comparison}
    \begin{tabular}{|c|c|c|c|c|c|c|c|}
    \hline
    \multirow{2}{*}{Method} & \multicolumn{3}{c|}{Novel-view Synthesis} & \multicolumn{3}{c|}{Relighting} & Shape \\
    \cline{2-8}
    & PSNR$\uparrow$ & SSIM$\uparrow$ & LPIPS$\downarrow$ & PSNR$\uparrow$ & SSIM$\uparrow$ & LPIPS$\downarrow$ & CD$\downarrow$\\
    \hline
    Ours & 30.70 & 0.97 & 0.019 & 27.99 & 0.96 & 0.040 & 8.35e-6 \\
    \hline
    Envmap & 22.41 & 0.95 & 0.054 & 21.32 & 0.92 & 0.088 & 1.20e-4 \\
    \hline
    \end{tabular}
    \end{table}

\begin{figure*}[t]
    \centering
        \begin{minipage}{0.03\textwidth}
            \centering
            \raisebox{\dimexpr0.5\baselineskip-\height}{\rotatebox{90}{{\bf Head}}}
        \end{minipage}%
        \begin{minipage}{0.78\textwidth}
            \includeinkscape[width=.98\linewidth]{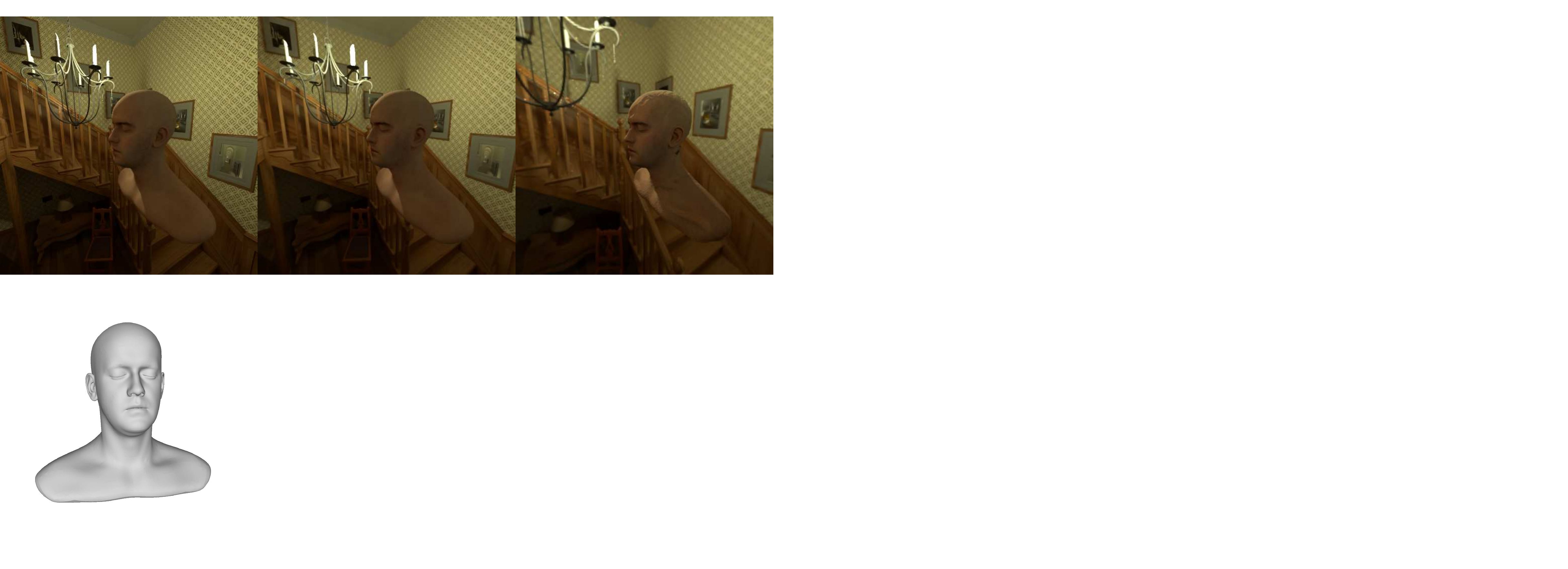}
        \end{minipage}
        \vspace{0.25em}
        \color{blue}
        \rule{.78\linewidth}{0.1pt} 
        \color{black}
        \vspace{0.25em}
        \begin{minipage}{0.03\textwidth}
            \centering
            \raisebox{\dimexpr0.5\baselineskip-\height}{\rotatebox{90}{{\bf Hotdog}}}
        \end{minipage}%
         \begin{minipage}{0.78\textwidth}
            \includeinkscape[width=.98\linewidth]{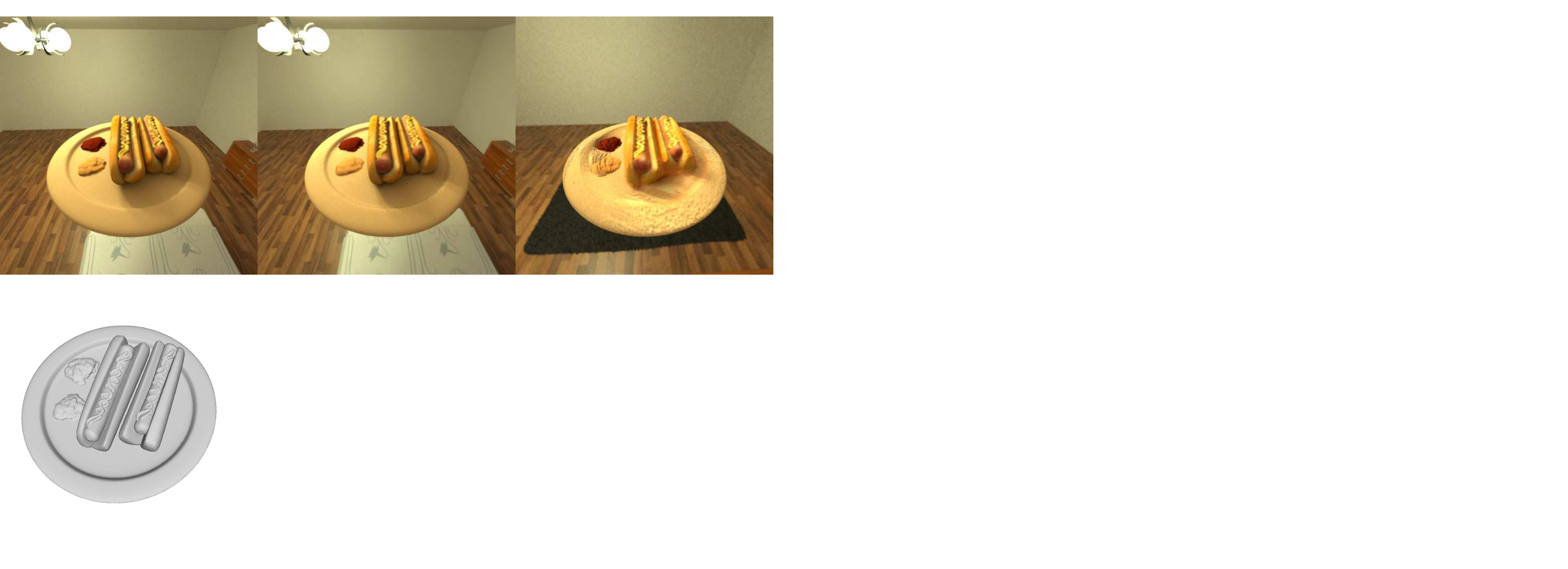}
        \end{minipage}
        \vspace{0.25em}
        \color{blue}
        \rule{.78\linewidth}{0.1pt} 
        \color{black}
        \vspace{0.25em}
        \begin{minipage}{0.03\textwidth}
            \centering
            \raisebox{\dimexpr0.5\baselineskip-\height}{\rotatebox{90}{{\bf Teapot}}}
        \end{minipage}%
        \begin{minipage}{0.78\textwidth}
            \includeinkscape[width=.98\linewidth]{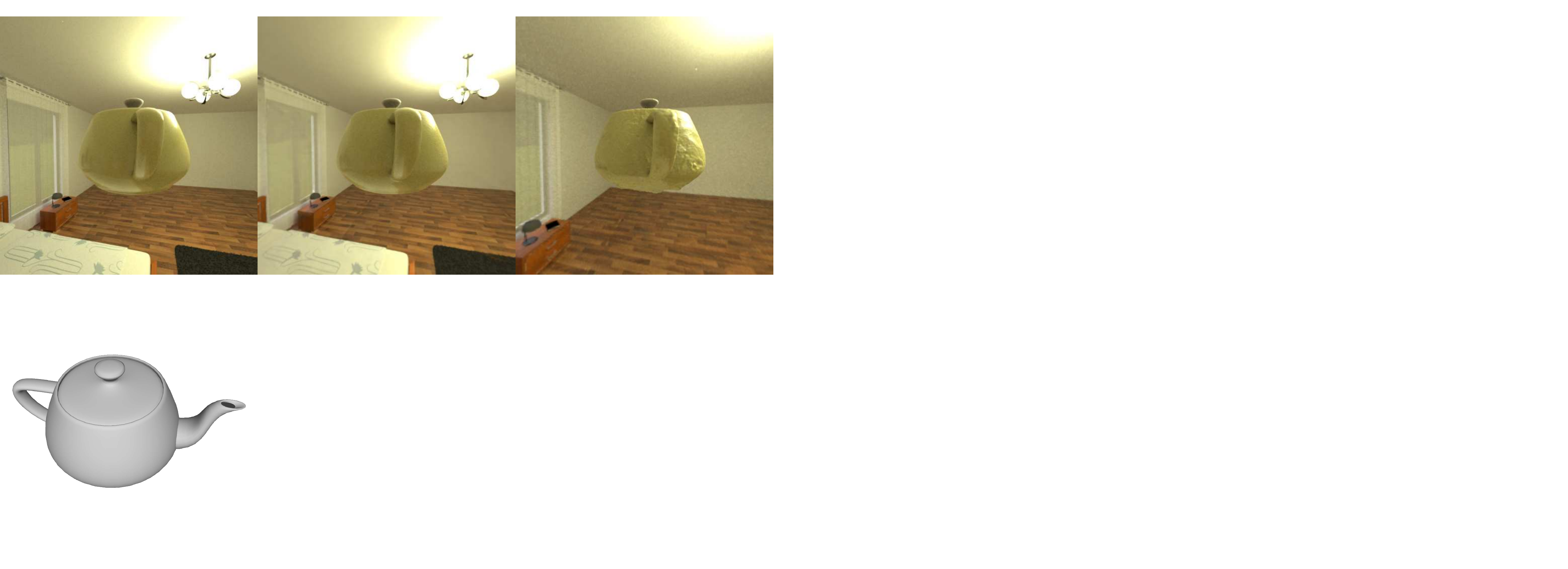}
        \end{minipage}
        \vspace{0.25em}
        \color{blue}
        \rule{.78\linewidth}{0.1pt} 
        \color{black}
        \vspace{0.25em}
        \begin{minipage}{0.03\textwidth}
            \centering
            \raisebox{\dimexpr0.5\baselineskip-\height}{\rotatebox{90}{\bf Boar}}
        \end{minipage}%
        \begin{minipage}{0.78\textwidth}
            \includeinkscape[width=.98\linewidth]{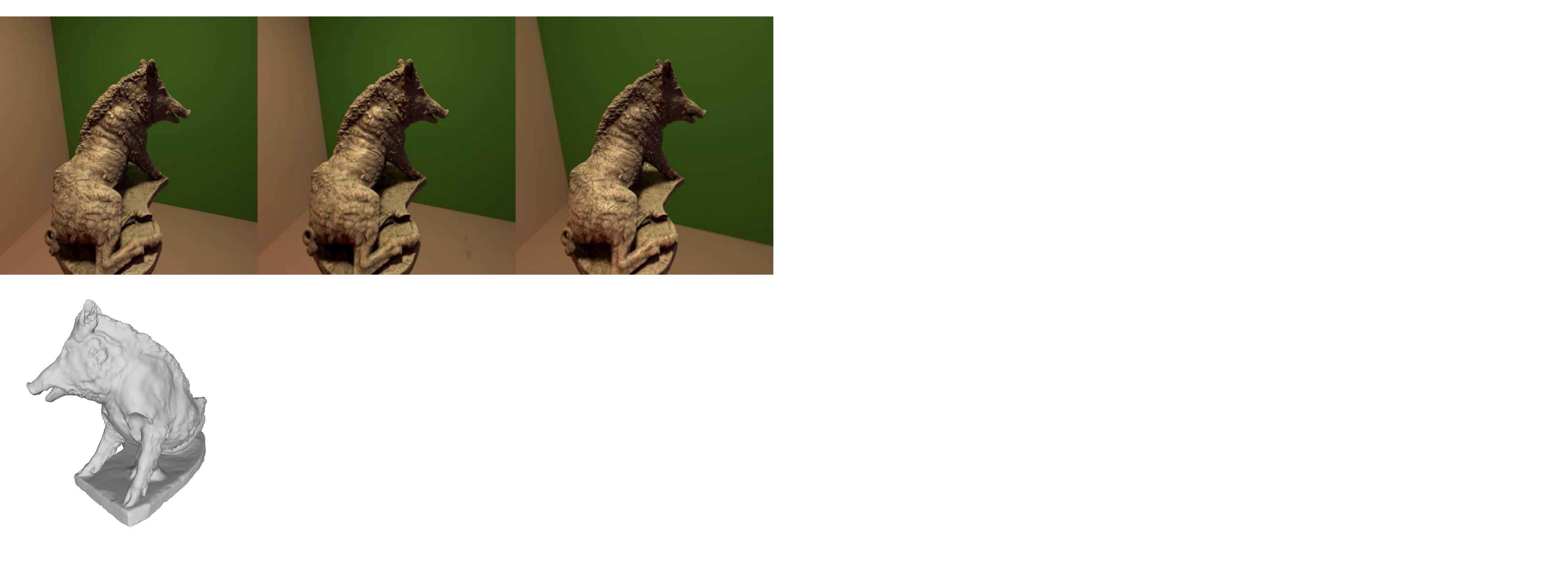}
        \end{minipage}
    \caption{Comparison with the environment map baseline on synthetic datasets.}
    \label{fig:syntheticone}
\end{figure*}

Figure~\ref{fig:syntheticone} presents the qualitative comparison results of the rendering and shape reconstruction between our method and the environment map baseline on the synthetic datasets.
We visualize rendered images under the original lighting, reconstructed shape, and relighting results in two novel lighting conditions.
For the Head and Hotdog scene, we optimize the diffuse reflectance parameter using a diffuse BSDF following the approach by \citet{DBLP:journals/tog/ViciniSJ22}.
In other scenes, we optimize the base color and roughness parameters using a principled BSDF \cite{Bur12}.
In the {\bf Head} scene, shadows cast by the head onto the shoulder and chest are evident.
While the baseline appears to match these cast shadows when rendering in the original lighting, the relighting results reveal local shape bumps.
These stem from inaccuracies during inverse rendering, where shadows rendered by the environment map fail to align with input images at shadow borders.
In the {\bf Hotdog} scene, inaccurate shadows rendered by the environment map compromise the rendering results under original lighting.
These artifacts persist during relighting and shape visualization.
Given our method's use of a more accurate light modeling approach using NeRF, rendered shadows better align with input images under original lighting, and our relighting and shape results do not exhibit such issues.
In the {\bf Teapot} scene, glossy highlights are apparent.
However, due to environment map inaccuracies, baseline-rendered highlights do not match reference images under original lighting, resulting in inaccurate material parameters and geometry bumps, visible in relighting results. In contrast, our method more closely replicates glossy highlights in both original lighting and relighting results.
The {\bf Boar} scene is captured in a Cornell box with a red wall on one side and a green wall on the other.
The baseline appears to show illumination effects from the red wall on the boar's hind leg in the original lighting.
However, relighting results reveal baked illumination effects (green on the boar's face, red on the boar's hind leg).
Our method accurately renders illumination effects under original lighting without baking reflections in relighting results.

In Table~\ref{tab:comparison}, we present the average quantitative results of our method and the environment map baseline on the synthetic dataset.
We use ground truth object masks to set background pixels to black when evaluating novel-view synthesis and relighting.
This approach provides a more precise representation of the object foreground's quality, excluding the impact of the inaccurate background in the baseline.
The PSNR metrics are exclusively computed within the object foreground mask.
Notably, our method consistently outperforms the baseline across novel-view synthesis, relighting, and shape reconstruction tasks.

\begin{figure*}[t]
    \centering
        \begin{minipage}{0.03\textwidth}
            \centering
            \raisebox{\dimexpr0.5\baselineskip-\height}{\rotatebox{90}{\bf Cabbage}}
        \end{minipage}%
        \begin{minipage}{0.9\textwidth}
            \includeinkscape[width=.98\linewidth]{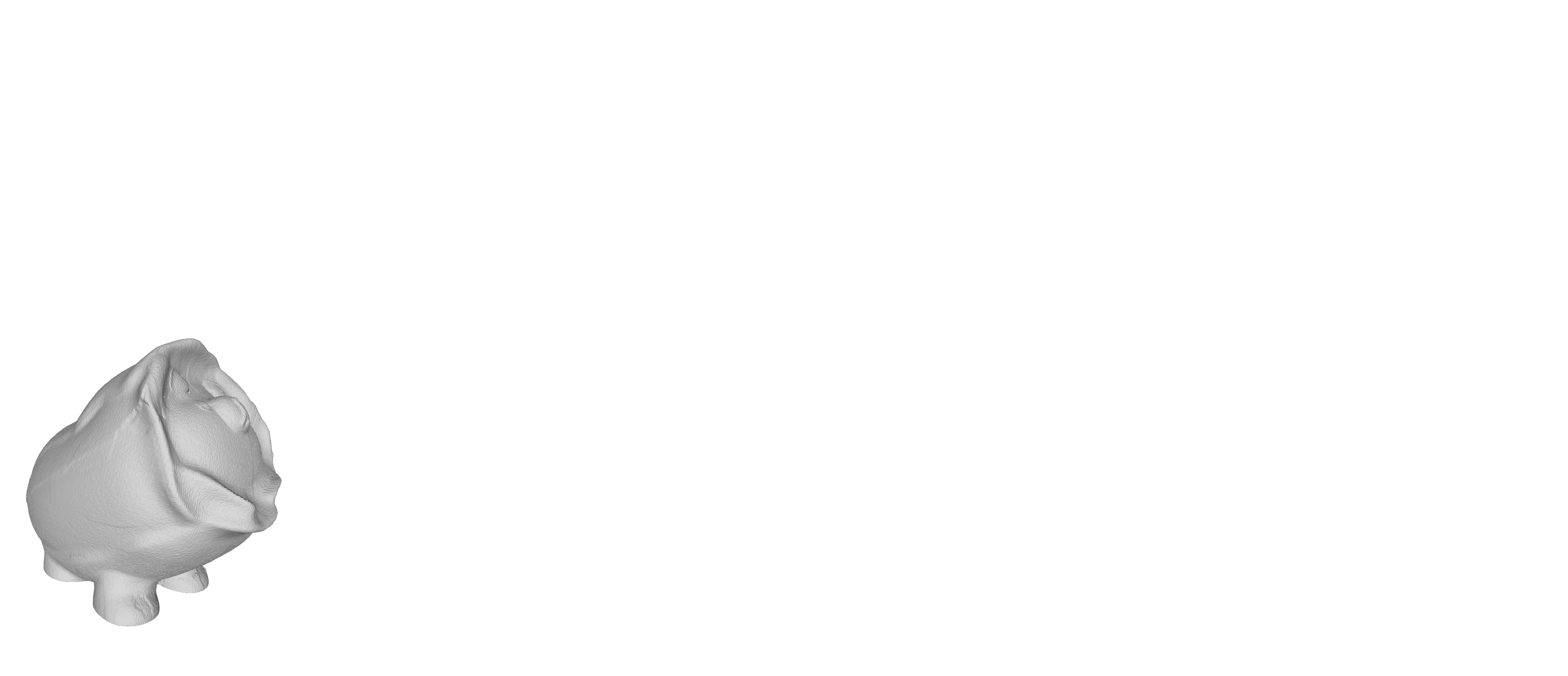}
        \end{minipage}
        \vspace{0.25em}
        \color{blue}
        \rule{.9\linewidth}{0.1pt} 
        \color{black}
        \vspace{0.25em}
        \begin{minipage}{0.03\textwidth}
            \centering
            \raisebox{\dimexpr0.5\baselineskip-\height}{\rotatebox{90}{\bf RealChair}}
        \end{minipage}%
        \begin{minipage}{0.9\textwidth}
            \includeinkscape[width=.98\linewidth]{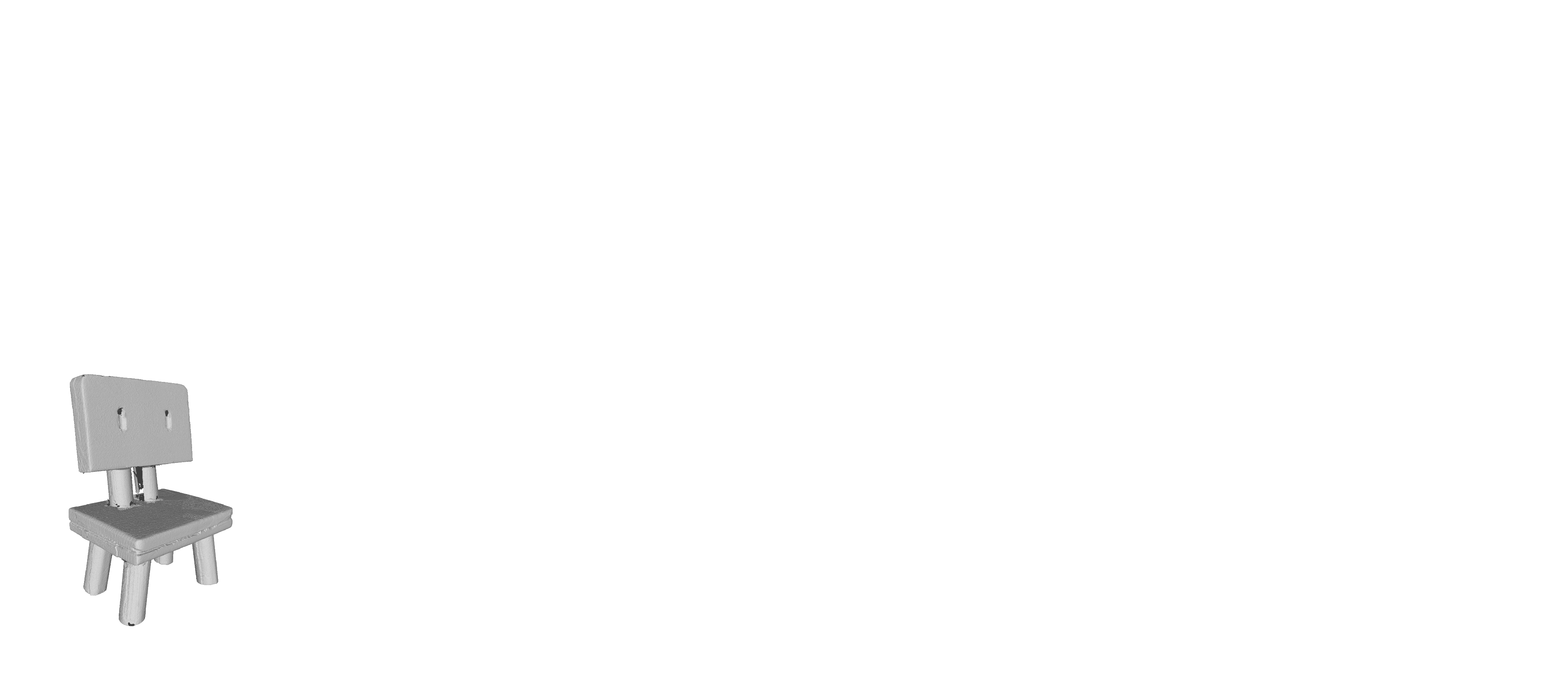}
        \end{minipage}
        \vspace{0.25em}
        \color{blue}
        \rule{.9\linewidth}{0.1pt} 
        \color{black}
        \vspace{0.25em}
        \begin{minipage}{0.03\textwidth}
            \centering
            \raisebox{\dimexpr0.5\baselineskip-\height}{\rotatebox{90}{\bf Dog}}
        \end{minipage}%
        \begin{minipage}{0.9\textwidth}
            \includeinkscape[width=.98\linewidth]{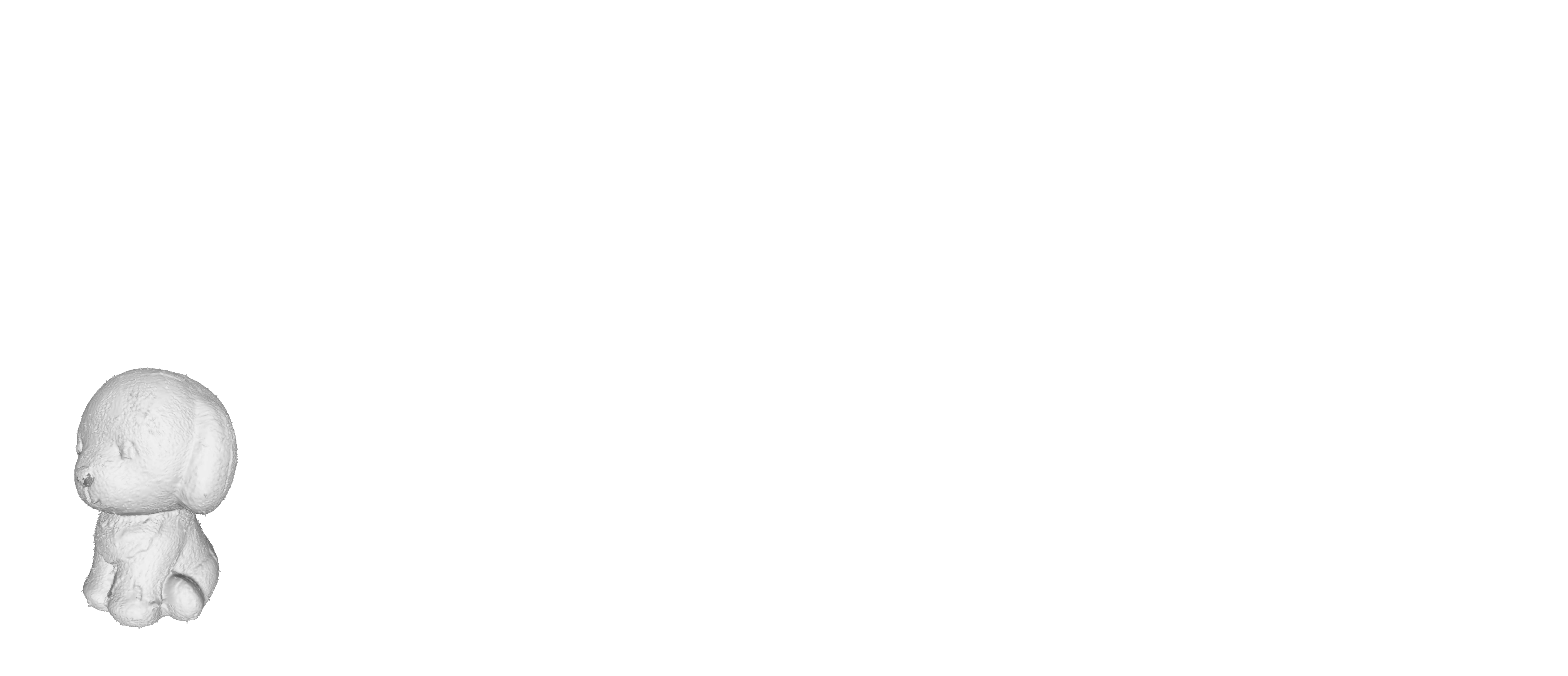}
        \end{minipage}
    \caption{Comparison with the environment map baseline on real datasets.}
    \label{fig:realone}
\end{figure*}

Figure~\ref{fig:realone}~and~\ref{fig:teaser} show the comparison results on real-world data.
In the {\bf Cabbage}, {\bf Hamster} and {\bf Dog} scenes, we observe attached shadows in the original lighting.
It is evident in the relighting results that many attached shadows are baked into the baseline, causing dark artifacts.
Our method, in contrast, produces more reasonable relighting results, indicating a more accurate material reconstruction.
The {\bf RealChair} scene exhibits strong self-shadows.
We observe inaccurately cast shadows rendered by the baseline in the original lighting.
These mismatched shadow borders are also present in the reconstructed materials during relighting. 
Our method accurately renders cast shadows in the original lighting and yields cleaner relighting results.
We visualize the shape errors as heatmaps in the insets. From these visualized shape and error maps, we observe that our reconstructed shape is more accurate than the baseline.
\rev{
A part of the checkerboard is enclosed within the bounding box, thus modeled by surface rendering, while the rest is modeled by NeRF. During shape visualization, we manually crop the checkerboard part to visualize the object only.
}

\subsection{Ablation Study}
We compare our emitter importance sampling for NeRF with pure BSDF sampling in Figure~\ref{fig:gradient_image}.

In the top row, we show two examples containing, respectively, a rough cube and a rough sphere with a NeRF emitter trained using multi-view renderings of the Cornell box.
The camera is positioned in a top-down perspective view. 
The derivative images are computed with respect to the vertical displacement of the cube and the radius of the sphere.

The bottom row of Figure~\ref{fig:gradient_image} show two examples under more complex non-distant indoor illuminations, both exhibiting self-shadowing effects.
We compute derivative images with respect to vertical displacements of the objects.

In all four examples, converged derivative estimates (using 8192 spp) generated with our approach closely match finite-difference references (using 65536 spp).
Using lower sample counts (256 spp), our importance sampling outperforms pure BSDF sampling significantly.

\section{Discussion and Conclusion}

\paragraph*{Discussion}
The main drawback of employing our proposed NeRF emitter is the increased computational cost in evaluating NeRF.
With our multi-GPU implementation, our inverse rendering optimization requires approximately 4.5 hours on 8 RTX 4090 GPUs, whereas the baseline using an environment map completes in about one hour on a single RTX 4090. 
However, accelerating NeRF remains an active area of research, and our method stands to benefit from advancements in this direction.
Recent works such as Adaptive Shells \cite{DBLP:journals/tog/WangSNSGKFMG23} can potentially accelerate our approach by improving NeRF evaluation speed.
Additionally, research progress aimed at reducing the samples per pixel in inverse rendering, exemplified by the works of  \citet{DBLP:journals/tog/NicoletRNKJM23,DBLP:journals/tog/WangWWZ23,DBLP:conf/siggraph/ChangSNHRL23}, may further accelerate our approach by reducing the number of NeRF evaluation rays.

Our emitter sampling method for NeRF does not account for the directional dependency of the radiance function $\radiance$.
While this assumption performs well with diffuse emitters, it can result in increased rendering variance with highly directional-dependent emitters
It also overlooks occlusion introduced by the surface to NeRF, which should decrease the sampling probabilities of occluded emitters.
However, as an initial step in the emitter importance sampling for NeRF emitters, our proposed method achieves more accurate light modeling and effective light sampling in inverse rendering.
Future work could explore this direction further to reduce the rendering variance when using NeRF as an emitter, possibly employing more advanced guiding techniques \cite{DBLP:journals/cgf/MullerGN17}.

\paragraph*{Conclusion}
Our work showcased the efficacy of integrating NeRF as a non-distant environment emitter within the physics-based inverse rendering pipeline. 
By addressing the shortcomings of the widely adopted distant environmental lighting model, we developed a novel approach to capture spatially varying illumination for accurate 3D reconstruction under real-world configurations. 
Furthermore, our introduction of an emitter importance sampling technique offered significant variance reduction for rendering with NeRF-based lighting.
We demonstrated the effectiveness of our technique using both real and synthetic inverse-rendering experiments.
\begin{acks}
    This work was supported by the National Key R\&D Program of China (2023YFC3305600, 2018YFA0704000), the NSFC (No.62021002), and the Key Research and Development Project of Tibet Autonomous Region (XZ202101ZY0019G). This work was also supported by THUIBCS, Tsinghua University, and BLBCI, Beijing Municipal Education Commission. Feng Xu is the corresponding author.
\end{acks}
\bibliographystyle{ACM-Reference-Format}
\bibliography{bibliography}


\begin{thebibliography}{61}


\ifx \showCODEN    \undefined \def \showCODEN     #1{\unskip}     \fi
\ifx \showDOI      \undefined \def \showDOI       #1{#1}\fi
\ifx \showISBNx    \undefined \def \showISBNx     #1{\unskip}     \fi
\ifx \showISBNxiii \undefined \def \showISBNxiii  #1{\unskip}     \fi
\ifx \showISSN     \undefined \def \showISSN      #1{\unskip}     \fi
\ifx \showLCCN     \undefined \def \showLCCN      #1{\unskip}     \fi
\ifx \shownote     \undefined \def \shownote      #1{#1}          \fi
\ifx \showarticletitle \undefined \def \showarticletitle #1{#1}   \fi
\ifx \showURL      \undefined \def \showURL       {\relax}        \fi
\providecommand\bibfield[2]{#2}
\providecommand\bibinfo[2]{#2}
\providecommand\natexlab[1]{#1}
\providecommand\showeprint[2][]{arXiv:#2}

\bibitem[Bangaru et~al\mbox{.}(2022)]%
        {DBLP:conf/siggrapha/BangaruGLLSHBXB22}
\bibfield{author}{\bibinfo{person}{Sai~Praveen Bangaru}, \bibinfo{person}{Micha{\"{e}}l Gharbi}, \bibinfo{person}{Fujun Luan}, \bibinfo{person}{Tzu{-}Mao Li}, \bibinfo{person}{Kalyan Sunkavalli}, \bibinfo{person}{Milos Hasan}, \bibinfo{person}{Sai Bi}, \bibinfo{person}{Zexiang Xu}, \bibinfo{person}{Gilbert Bernstein}, {and} \bibinfo{person}{Fr{\'{e}}do Durand}.} \bibinfo{year}{2022}\natexlab{}.
\newblock \showarticletitle{Differentiable Rendering of Neural SDFs through Reparameterization}. In \bibinfo{booktitle}{\emph{{SIGGRAPH} Asia 2022 Conference Papers, {SA} 2022, Daegu, Republic of Korea, December 6-9, 2022}}, \bibfield{editor}{\bibinfo{person}{Soon~Ki Jung}, \bibinfo{person}{Jehee Lee}, {and} \bibinfo{person}{Adam~W. Bargteil}} (Eds.). \bibinfo{publisher}{{ACM}}, \bibinfo{pages}{5:1--5:9}.
\newblock
\urldef\tempurl%
\url{https://doi.org/10.1145/3550469.3555397}
\showDOI{\tempurl}


\bibitem[Bangaru et~al\mbox{.}(2020)]%
        {DBLP:journals/tog/BangaruLD20}
\bibfield{author}{\bibinfo{person}{Sai~Praveen Bangaru}, \bibinfo{person}{Tzu{-}Mao Li}, {and} \bibinfo{person}{Fr{\'{e}}do Durand}.} \bibinfo{year}{2020}\natexlab{}.
\newblock \showarticletitle{Unbiased warped-area sampling for differentiable rendering}.
\newblock \bibinfo{journal}{\emph{{ACM} Trans. Graph.}} \bibinfo{volume}{39}, \bibinfo{number}{6} (\bibinfo{year}{2020}), \bibinfo{pages}{245:1--245:18}.
\newblock
\urldef\tempurl%
\url{https://doi.org/10.1145/3414685.3417833}
\showDOI{\tempurl}


\bibitem[Barron et~al\mbox{.}(2022)]%
        {DBLP:conf/cvpr/BarronMVSH22}
\bibfield{author}{\bibinfo{person}{Jonathan~T. Barron}, \bibinfo{person}{Ben Mildenhall}, \bibinfo{person}{Dor Verbin}, \bibinfo{person}{Pratul~P. Srinivasan}, {and} \bibinfo{person}{Peter Hedman}.} \bibinfo{year}{2022}\natexlab{}.
\newblock \showarticletitle{Mip-NeRF 360: Unbounded Anti-Aliased Neural Radiance Fields}. In \bibinfo{booktitle}{\emph{{IEEE/CVF} Conference on Computer Vision and Pattern Recognition, {CVPR} 2022, New Orleans, LA, USA, June 18-24, 2022}}. \bibinfo{publisher}{{IEEE}}, \bibinfo{pages}{5460--5469}.
\newblock
\urldef\tempurl%
\url{https://doi.org/10.1109/CVPR52688.2022.00539}
\showDOI{\tempurl}


\bibitem[Barron et~al\mbox{.}(2023)]%
        {DBLP:conf/iccv/BarronMVSH23}
\bibfield{author}{\bibinfo{person}{Jonathan~T. Barron}, \bibinfo{person}{Ben Mildenhall}, \bibinfo{person}{Dor Verbin}, \bibinfo{person}{Pratul~P. Srinivasan}, {and} \bibinfo{person}{Peter Hedman}.} \bibinfo{year}{2023}\natexlab{}.
\newblock \showarticletitle{Zip-NeRF: Anti-Aliased Grid-Based Neural Radiance Fields}. In \bibinfo{booktitle}{\emph{{IEEE/CVF} International Conference on Computer Vision, {ICCV} 2023, Paris, France, October 1-6, 2023}}. \bibinfo{publisher}{{IEEE}}, \bibinfo{pages}{19640--19648}.
\newblock
\urldef\tempurl%
\url{https://doi.org/10.1109/ICCV51070.2023.01804}
\showDOI{\tempurl}


\bibitem[Bi et~al\mbox{.}(2020a)]%
        {DBLP:journals/corr/abs-2008-03824}
\bibfield{author}{\bibinfo{person}{Sai Bi}, \bibinfo{person}{Zexiang Xu}, \bibinfo{person}{Pratul~P. Srinivasan}, \bibinfo{person}{Ben Mildenhall}, \bibinfo{person}{Kalyan Sunkavalli}, \bibinfo{person}{Milos Hasan}, \bibinfo{person}{Yannick Hold{-}Geoffroy}, \bibinfo{person}{David~J. Kriegman}, {and} \bibinfo{person}{Ravi Ramamoorthi}.} \bibinfo{year}{2020}\natexlab{a}.
\newblock \showarticletitle{Neural Reflectance Fields for Appearance Acquisition}.
\newblock \bibinfo{journal}{\emph{CoRR}}  \bibinfo{volume}{abs/2008.03824} (\bibinfo{year}{2020}).
\newblock
\showeprint[arXiv]{2008.03824}
\urldef\tempurl%
\url{https://arxiv.org/abs/2008.03824}
\showURL{%
\tempurl}


\bibitem[Bi et~al\mbox{.}(2020b)]%
        {DBLP:conf/eccv/BiXSHHKR20}
\bibfield{author}{\bibinfo{person}{Sai Bi}, \bibinfo{person}{Zexiang Xu}, \bibinfo{person}{Kalyan Sunkavalli}, \bibinfo{person}{Milos Hasan}, \bibinfo{person}{Yannick Hold{-}Geoffroy}, \bibinfo{person}{David~J. Kriegman}, {and} \bibinfo{person}{Ravi Ramamoorthi}.} \bibinfo{year}{2020}\natexlab{b}.
\newblock \showarticletitle{Deep Reflectance Volumes: Relightable Reconstructions from Multi-view Photometric Images}. In \bibinfo{booktitle}{\emph{Computer Vision - {ECCV} 2020 - 16th European Conference, Glasgow, UK, August 23-28, 2020, Proceedings, Part {III}}} \emph{(\bibinfo{series}{Lecture Notes in Computer Science}, Vol.~\bibinfo{volume}{12348})}, \bibfield{editor}{\bibinfo{person}{Andrea Vedaldi}, \bibinfo{person}{Horst Bischof}, \bibinfo{person}{Thomas Brox}, {and} \bibinfo{person}{Jan{-}Michael Frahm}} (Eds.). \bibinfo{publisher}{Springer}, \bibinfo{pages}{294--311}.
\newblock
\urldef\tempurl%
\url{https://doi.org/10.1007/978-3-030-58580-8\_18}
\showDOI{\tempurl}


\bibitem[Boss et~al\mbox{.}(2021a)]%
        {DBLP:conf/iccv/BossBJBLL21}
\bibfield{author}{\bibinfo{person}{Mark Boss}, \bibinfo{person}{Raphael Braun}, \bibinfo{person}{Varun Jampani}, \bibinfo{person}{Jonathan~T. Barron}, \bibinfo{person}{Ce Liu}, {and} \bibinfo{person}{Hendrik P.~A. Lensch}.} \bibinfo{year}{2021}\natexlab{a}.
\newblock \showarticletitle{NeRD: Neural Reflectance Decomposition from Image Collections}. In \bibinfo{booktitle}{\emph{2021 {IEEE/CVF} International Conference on Computer Vision, {ICCV} 2021, Montreal, QC, Canada, October 10-17, 2021}}. \bibinfo{publisher}{{IEEE}}, \bibinfo{pages}{12664--12674}.
\newblock
\urldef\tempurl%
\url{https://doi.org/10.1109/ICCV48922.2021.01245}
\showDOI{\tempurl}


\bibitem[Boss et~al\mbox{.}(2022)]%
        {DBLP:conf/nips/BossEKLSBLJ22}
\bibfield{author}{\bibinfo{person}{Mark Boss}, \bibinfo{person}{Andreas Engelhardt}, \bibinfo{person}{Abhishek Kar}, \bibinfo{person}{Yuanzhen Li}, \bibinfo{person}{Deqing Sun}, \bibinfo{person}{Jonathan~T. Barron}, \bibinfo{person}{Hendrik P.~A. Lensch}, {and} \bibinfo{person}{Varun Jampani}.} \bibinfo{year}{2022}\natexlab{}.
\newblock \showarticletitle{{SAMURAI:} Shape And Material from Unconstrained Real-world Arbitrary Image collections}. In \bibinfo{booktitle}{\emph{Advances in Neural Information Processing Systems 35: Annual Conference on Neural Information Processing Systems 2022, NeurIPS 2022, New Orleans, LA, USA, November 28 - December 9, 2022}}, \bibfield{editor}{\bibinfo{person}{Sanmi Koyejo}, \bibinfo{person}{S.~Mohamed}, \bibinfo{person}{A.~Agarwal}, \bibinfo{person}{Danielle Belgrave}, \bibinfo{person}{K.~Cho}, {and} \bibinfo{person}{A.~Oh}} (Eds.).
\newblock
\urldef\tempurl%
\url{http://papers.nips.cc/paper\_files/paper/2022/hash/a8f2713b5c6bdcd3d264f1aa9b9c6f03-Abstract-Conference.html}
\showURL{%
\tempurl}


\bibitem[Boss et~al\mbox{.}(2021b)]%
        {DBLP:conf/nips/BossJBLBL21}
\bibfield{author}{\bibinfo{person}{Mark Boss}, \bibinfo{person}{Varun Jampani}, \bibinfo{person}{Raphael Braun}, \bibinfo{person}{Ce Liu}, \bibinfo{person}{Jonathan~T. Barron}, {and} \bibinfo{person}{Hendrik P.~A. Lensch}.} \bibinfo{year}{2021}\natexlab{b}.
\newblock \showarticletitle{Neural-PIL: Neural Pre-Integrated Lighting for Reflectance Decomposition}. In \bibinfo{booktitle}{\emph{Advances in Neural Information Processing Systems 34: Annual Conference on Neural Information Processing Systems 2021, NeurIPS 2021, December 6-14, 2021, virtual}}, \bibfield{editor}{\bibinfo{person}{Marc'Aurelio Ranzato}, \bibinfo{person}{Alina Beygelzimer}, \bibinfo{person}{Yann~N. Dauphin}, \bibinfo{person}{Percy Liang}, {and} \bibinfo{person}{Jennifer~Wortman Vaughan}} (Eds.). \bibinfo{pages}{10691--10704}.
\newblock
\urldef\tempurl%
\url{https://proceedings.neurips.cc/paper/2021/hash/58ae749f25eded36f486bc85feb3f0ab-Abstract.html}
\showURL{%
\tempurl}


\bibitem[Burley(2012)]%
        {Bur12}
\bibfield{author}{\bibinfo{person}{Brent Burley}.} \bibinfo{year}{2012}\natexlab{}.
\newblock \showarticletitle{Physically Based Shading at Disney}. In \bibinfo{booktitle}{\emph{SIGGRAPH 2012 Course Notes}}.
\newblock
\urldef\tempurl%
\url{https://www.disneyanimation.com/publications/physically-based-shading-at-disney/}
\showURL{%
\tempurl}


\bibitem[Chang et~al\mbox{.}(2023)]%
        {DBLP:conf/siggraph/ChangSNHRL23}
\bibfield{author}{\bibinfo{person}{Wesley Chang}, \bibinfo{person}{Venkataram Sivaram}, \bibinfo{person}{Derek Nowrouzezahrai}, \bibinfo{person}{Toshiya Hachisuka}, \bibinfo{person}{Ravi Ramamoorthi}, {and} \bibinfo{person}{Tzu{-}Mao Li}.} \bibinfo{year}{2023}\natexlab{}.
\newblock \showarticletitle{Parameter-space ReSTIR for Differentiable and Inverse Rendering}. In \bibinfo{booktitle}{\emph{{ACM} {SIGGRAPH} 2023 Conference Proceedings, {SIGGRAPH} 2023, Los Angeles, CA, USA, August 6-10, 2023}}, \bibfield{editor}{\bibinfo{person}{Erik Brunvand}, \bibinfo{person}{Alla Sheffer}, {and} \bibinfo{person}{Michael Wimmer}} (Eds.). \bibinfo{publisher}{{ACM}}, \bibinfo{pages}{18:1--18:10}.
\newblock
\urldef\tempurl%
\url{https://doi.org/10.1145/3588432.3591512}
\showDOI{\tempurl}


\bibitem[Chen and Liu(2022)]%
        {DBLP:conf/eccv/ChenL22}
\bibfield{author}{\bibinfo{person}{Zhaoxi Chen} {and} \bibinfo{person}{Ziwei Liu}.} \bibinfo{year}{2022}\natexlab{}.
\newblock \showarticletitle{Relighting4D: Neural Relightable Human from Videos}. In \bibinfo{booktitle}{\emph{Computer Vision - {ECCV} 2022 - 17th European Conference, Tel Aviv, Israel, October 23-27, 2022, Proceedings, Part {XIV}}} \emph{(\bibinfo{series}{Lecture Notes in Computer Science}, Vol.~\bibinfo{volume}{13674})}, \bibfield{editor}{\bibinfo{person}{Shai Avidan}, \bibinfo{person}{Gabriel~J. Brostow}, \bibinfo{person}{Moustapha Ciss{\'{e}}}, \bibinfo{person}{Giovanni~Maria Farinella}, {and} \bibinfo{person}{Tal Hassner}} (Eds.). \bibinfo{publisher}{Springer}, \bibinfo{pages}{606--623}.
\newblock
\urldef\tempurl%
\url{https://doi.org/10.1007/978-3-031-19781-9\_35}
\showDOI{\tempurl}


\bibitem[Condor and Jarabo(2022)]%
        {DBLP:conf/rt/CondorJ22}
\bibfield{author}{\bibinfo{person}{Jorge Condor} {and} \bibinfo{person}{Adri{\'{a}}n Jarabo}.} \bibinfo{year}{2022}\natexlab{}.
\newblock \showarticletitle{A Learned Radiance-Field Representation for Complex Luminaires}. In \bibinfo{booktitle}{\emph{33rd Eurographics Symposium on Rendering, {EGSR} 2022 - Symposium Track, Prague, Czech Republic, 4-6 July 2022}}, \bibfield{editor}{\bibinfo{person}{Abhijeet Ghosh} {and} \bibinfo{person}{Li{-}Yi Wei}} (Eds.). \bibinfo{publisher}{Eurographics Association}, \bibinfo{pages}{49--58}.
\newblock
\urldef\tempurl%
\url{https://doi.org/10.2312/SR.20221155}
\showDOI{\tempurl}


\bibitem[Curless and Levoy(1996)]%
        {DBLP:conf/siggraph/CurlessL96}
\bibfield{author}{\bibinfo{person}{Brian Curless} {and} \bibinfo{person}{Marc Levoy}.} \bibinfo{year}{1996}\natexlab{}.
\newblock \showarticletitle{A Volumetric Method for Building Complex Models from Range Images}. In \bibinfo{booktitle}{\emph{Proceedings of the 23rd Annual Conference on Computer Graphics and Interactive Techniques, {SIGGRAPH} 1996, New Orleans, LA, USA, August 4-9, 1996}}, \bibfield{editor}{\bibinfo{person}{John Fujii}} (Ed.). \bibinfo{publisher}{{ACM}}, \bibinfo{pages}{303--312}.
\newblock
\urldef\tempurl%
\url{https://doi.org/10.1145/237170.237269}
\showDOI{\tempurl}


\bibitem[Debevec and Malik(1997)]%
        {DBLP:conf/siggraph/DebevecM97}
\bibfield{author}{\bibinfo{person}{Paul~E. Debevec} {and} \bibinfo{person}{Jitendra Malik}.} \bibinfo{year}{1997}\natexlab{}.
\newblock \showarticletitle{Recovering high dynamic range radiance maps from photographs}. In \bibinfo{booktitle}{\emph{Proceedings of the 24th Annual Conference on Computer Graphics and Interactive Techniques, {SIGGRAPH} 1997, Los Angeles, CA, USA, August 3-8, 1997}}, \bibfield{editor}{\bibinfo{person}{G.~Scott Owen}, \bibinfo{person}{Turner Whitted}, {and} \bibinfo{person}{Barbara Mones{-}Hattal}} (Eds.). \bibinfo{publisher}{{ACM}}, \bibinfo{pages}{369--378}.
\newblock
\urldef\tempurl%
\url{https://doi.org/10.1145/258734.258884}
\showDOI{\tempurl}


\bibitem[Ge et~al\mbox{.}(2023)]%
        {DBLP:conf/iccv/GeHZ0C23}
\bibfield{author}{\bibinfo{person}{Wenhang Ge}, \bibinfo{person}{Tao Hu}, \bibinfo{person}{Haoyu Zhao}, \bibinfo{person}{Shu Liu}, {and} \bibinfo{person}{Ying{-}Cong Chen}.} \bibinfo{year}{2023}\natexlab{}.
\newblock \showarticletitle{Ref-NeuS: Ambiguity-Reduced Neural Implicit Surface Learning for Multi-View Reconstruction with Reflection}. In \bibinfo{booktitle}{\emph{{IEEE/CVF} International Conference on Computer Vision, {ICCV} 2023, Paris, France, October 1-6, 2023}}. \bibinfo{publisher}{{IEEE}}, \bibinfo{pages}{4228--4237}.
\newblock
\urldef\tempurl%
\url{https://doi.org/10.1109/ICCV51070.2023.00392}
\showDOI{\tempurl}


\bibitem[Hanji and Mantiuk(2023)]%
        {DBLP:journals/tci/HanjiM23}
\bibfield{author}{\bibinfo{person}{Param Hanji} {and} \bibinfo{person}{Rafal~K. Mantiuk}.} \bibinfo{year}{2023}\natexlab{}.
\newblock \showarticletitle{Robust Estimation of Exposure Ratios in Multi-Exposure Image Stacks}.
\newblock \bibinfo{journal}{\emph{{IEEE} Trans. Computational Imaging}}  \bibinfo{volume}{9} (\bibinfo{year}{2023}), \bibinfo{pages}{721--731}.
\newblock
\urldef\tempurl%
\url{https://doi.org/10.1109/TCI.2023.3301338}
\showDOI{\tempurl}


\bibitem[Hanji et~al\mbox{.}(2020)]%
        {DBLP:conf/eccv/HanjiZM20}
\bibfield{author}{\bibinfo{person}{Param Hanji}, \bibinfo{person}{Fangcheng Zhong}, {and} \bibinfo{person}{Rafal~K. Mantiuk}.} \bibinfo{year}{2020}\natexlab{}.
\newblock \showarticletitle{Noise-Aware Merging of High Dynamic Range Image Stacks Without Camera Calibration}. In \bibinfo{booktitle}{\emph{Computer Vision - {ECCV} 2020 Workshops - Glasgow, UK, August 23-28, 2020, Proceedings, Part {III}}} \emph{(\bibinfo{series}{Lecture Notes in Computer Science}, Vol.~\bibinfo{volume}{12537})}, \bibfield{editor}{\bibinfo{person}{Adrien Bartoli} {and} \bibinfo{person}{Andrea Fusiello}} (Eds.). \bibinfo{publisher}{Springer}, \bibinfo{pages}{376--391}.
\newblock
\urldef\tempurl%
\url{https://doi.org/10.1007/978-3-030-67070-2\_23}
\showDOI{\tempurl}


\bibitem[Hasselgren et~al\mbox{.}(2022)]%
        {DBLP:conf/nips/HasselgrenHM22}
\bibfield{author}{\bibinfo{person}{Jon Hasselgren}, \bibinfo{person}{Nikolai Hofmann}, {and} \bibinfo{person}{Jacob Munkberg}.} \bibinfo{year}{2022}\natexlab{}.
\newblock \showarticletitle{Shape, Light, and Material Decomposition from Images using Monte Carlo Rendering and Denoising}. In \bibinfo{booktitle}{\emph{Advances in Neural Information Processing Systems 35: Annual Conference on Neural Information Processing Systems 2022, NeurIPS 2022, New Orleans, LA, USA, November 28 - December 9, 2022}}, \bibfield{editor}{\bibinfo{person}{Sanmi Koyejo}, \bibinfo{person}{S.~Mohamed}, \bibinfo{person}{A.~Agarwal}, \bibinfo{person}{Danielle Belgrave}, \bibinfo{person}{K.~Cho}, {and} \bibinfo{person}{A.~Oh}} (Eds.).
\newblock
\urldef\tempurl%
\url{http://papers.nips.cc/paper\_files/paper/2022/hash/8fcb27984bf16ca03cad643244ec470d-Abstract-Conference.html}
\showURL{%
\tempurl}


\bibitem[Jakob(2012)]%
        {jakob2012numerically}
\bibfield{author}{\bibinfo{person}{Wenzel Jakob}.} \bibinfo{year}{2012}\natexlab{}.
\newblock \showarticletitle{Numerically stable sampling of the von Mises-Fisher distribution on Sˆ2 (and other tricks)}.
\newblock \bibinfo{journal}{\emph{Interactive Geometry Lab, ETH Z{\"u}rich, Tech. Rep}} (\bibinfo{year}{2012}), \bibinfo{pages}{6}.
\newblock


\bibitem[Jakob et~al\mbox{.}(2022b)]%
        {jakob2022mitsuba3}
\bibfield{author}{\bibinfo{person}{Wenzel Jakob}, \bibinfo{person}{Sébastien Speierer}, \bibinfo{person}{Nicolas Roussel}, \bibinfo{person}{Merlin Nimier-David}, \bibinfo{person}{Delio Vicini}, \bibinfo{person}{Tizian Zeltner}, \bibinfo{person}{Baptiste Nicolet}, \bibinfo{person}{Miguel Crespo}, \bibinfo{person}{Vincent Leroy}, {and} \bibinfo{person}{Ziyi Zhang}.} \bibinfo{year}{2022}\natexlab{b}.
\newblock \bibinfo{booktitle}{\emph{Mitsuba 3 renderer}}.
\newblock
\newblock
\shownote{https://mitsuba-renderer.org}.


\bibitem[Jakob et~al\mbox{.}(2022a)]%
        {DBLP:journals/tog/JakobSRV22}
\bibfield{author}{\bibinfo{person}{Wenzel Jakob}, \bibinfo{person}{S{\'{e}}bastien Speierer}, \bibinfo{person}{Nicolas Roussel}, {and} \bibinfo{person}{Delio Vicini}.} \bibinfo{year}{2022}\natexlab{a}.
\newblock \showarticletitle{{DR.JIT:} a just-in-time compiler for differentiable rendering}.
\newblock \bibinfo{journal}{\emph{{ACM} Trans. Graph.}} \bibinfo{volume}{41}, \bibinfo{number}{4} (\bibinfo{year}{2022}), \bibinfo{pages}{124:1--124:19}.
\newblock
\urldef\tempurl%
\url{https://doi.org/10.1145/3528223.3530099}
\showDOI{\tempurl}


\bibitem[Jin et~al\mbox{.}(2023)]%
        {DBLP:conf/cvpr/JinLXZHBZX023}
\bibfield{author}{\bibinfo{person}{Haian Jin}, \bibinfo{person}{Isabella Liu}, \bibinfo{person}{Peijia Xu}, \bibinfo{person}{Xiaoshuai Zhang}, \bibinfo{person}{Songfang Han}, \bibinfo{person}{Sai Bi}, \bibinfo{person}{Xiaowei Zhou}, \bibinfo{person}{Zexiang Xu}, {and} \bibinfo{person}{Hao Su}.} \bibinfo{year}{2023}\natexlab{}.
\newblock \showarticletitle{TensoIR: Tensorial Inverse Rendering}. In \bibinfo{booktitle}{\emph{{IEEE/CVF} Conference on Computer Vision and Pattern Recognition, {CVPR} 2023, Vancouver, BC, Canada, June 17-24, 2023}}. \bibinfo{publisher}{{IEEE}}, \bibinfo{pages}{165--174}.
\newblock
\urldef\tempurl%
\url{https://doi.org/10.1109/CVPR52729.2023.00024}
\showDOI{\tempurl}


\bibitem[Karl{\'{\i}}k et~al\mbox{.}(2019)]%
        {DBLP:journals/tog/KarlikSVSK19}
\bibfield{author}{\bibinfo{person}{Ondrej Karl{\'{\i}}k}, \bibinfo{person}{Martin Sik}, \bibinfo{person}{Petr V{\'{e}}voda}, \bibinfo{person}{Tom{\'{a}}s Skrivan}, {and} \bibinfo{person}{Jaroslav Kriv{\'{a}}nek}.} \bibinfo{year}{2019}\natexlab{}.
\newblock \showarticletitle{{MIS} compensation: optimizing sampling techniques in multiple importance sampling}.
\newblock \bibinfo{journal}{\emph{{ACM} Trans. Graph.}} \bibinfo{volume}{38}, \bibinfo{number}{6} (\bibinfo{year}{2019}), \bibinfo{pages}{151:1--151:12}.
\newblock
\urldef\tempurl%
\url{https://doi.org/10.1145/3355089.3356565}
\showDOI{\tempurl}


\bibitem[Kingma and Ba(2015)]%
        {DBLP:journals/corr/KingmaB14}
\bibfield{author}{\bibinfo{person}{Diederik~P. Kingma} {and} \bibinfo{person}{Jimmy Ba}.} \bibinfo{year}{2015}\natexlab{}.
\newblock \showarticletitle{Adam: {A} Method for Stochastic Optimization}. In \bibinfo{booktitle}{\emph{3rd International Conference on Learning Representations, {ICLR} 2015, San Diego, CA, USA, May 7-9, 2015, Conference Track Proceedings}}, \bibfield{editor}{\bibinfo{person}{Yoshua Bengio} {and} \bibinfo{person}{Yann LeCun}} (Eds.).
\newblock
\urldef\tempurl%
\url{http://arxiv.org/abs/1412.6980}
\showURL{%
\tempurl}


\bibitem[Li et~al\mbox{.}(2018)]%
        {DBLP:journals/tog/LiADL18}
\bibfield{author}{\bibinfo{person}{Tzu{-}Mao Li}, \bibinfo{person}{Miika Aittala}, \bibinfo{person}{Fr{\'{e}}do Durand}, {and} \bibinfo{person}{Jaakko Lehtinen}.} \bibinfo{year}{2018}\natexlab{}.
\newblock \showarticletitle{Differentiable Monte Carlo ray tracing through edge sampling}.
\newblock \bibinfo{journal}{\emph{{ACM} Trans. Graph.}} \bibinfo{volume}{37}, \bibinfo{number}{6} (\bibinfo{year}{2018}), \bibinfo{pages}{222}.
\newblock
\urldef\tempurl%
\url{https://doi.org/10.1145/3272127.3275109}
\showDOI{\tempurl}


\bibitem[Lin et~al\mbox{.}(2021)]%
        {DBLP:conf/iccv/LinM0L21}
\bibfield{author}{\bibinfo{person}{Chen{-}Hsuan Lin}, \bibinfo{person}{Wei{-}Chiu Ma}, \bibinfo{person}{Antonio Torralba}, {and} \bibinfo{person}{Simon Lucey}.} \bibinfo{year}{2021}\natexlab{}.
\newblock \showarticletitle{{BARF:} Bundle-Adjusting Neural Radiance Fields}. In \bibinfo{booktitle}{\emph{2021 {IEEE/CVF} International Conference on Computer Vision, {ICCV} 2021, Montreal, QC, Canada, October 10-17, 2021}}. \bibinfo{publisher}{{IEEE}}, \bibinfo{pages}{5721--5731}.
\newblock
\urldef\tempurl%
\url{https://doi.org/10.1109/ICCV48922.2021.00569}
\showDOI{\tempurl}


\bibitem[Ling et~al\mbox{.}(2023)]%
        {DBLP:conf/cvpr/Ling0023}
\bibfield{author}{\bibinfo{person}{Jingwang Ling}, \bibinfo{person}{Zhibo Wang}, {and} \bibinfo{person}{Feng Xu}.} \bibinfo{year}{2023}\natexlab{}.
\newblock \showarticletitle{ShadowNeuS: Neural {SDF} Reconstruction by Shadow Ray Supervision}. In \bibinfo{booktitle}{\emph{{IEEE/CVF} Conference on Computer Vision and Pattern Recognition, {CVPR} 2023, Vancouver, BC, Canada, June 17-24, 2023}}. \bibinfo{publisher}{{IEEE}}, \bibinfo{pages}{175--185}.
\newblock
\urldef\tempurl%
\url{https://doi.org/10.1109/CVPR52729.2023.00025}
\showDOI{\tempurl}


\bibitem[Mildenhall et~al\mbox{.}(2022)]%
        {DBLP:conf/cvpr/MildenhallHMSB22}
\bibfield{author}{\bibinfo{person}{Ben Mildenhall}, \bibinfo{person}{Peter Hedman}, \bibinfo{person}{Ricardo Martin{-}Brualla}, \bibinfo{person}{Pratul~P. Srinivasan}, {and} \bibinfo{person}{Jonathan~T. Barron}.} \bibinfo{year}{2022}\natexlab{}.
\newblock \showarticletitle{NeRF in the Dark: High Dynamic Range View Synthesis from Noisy Raw Images}. In \bibinfo{booktitle}{\emph{{IEEE/CVF} Conference on Computer Vision and Pattern Recognition, {CVPR} 2022, New Orleans, LA, USA, June 18-24, 2022}}. \bibinfo{publisher}{{IEEE}}, \bibinfo{pages}{16169--16178}.
\newblock
\urldef\tempurl%
\url{https://doi.org/10.1109/CVPR52688.2022.01571}
\showDOI{\tempurl}


\bibitem[Mildenhall et~al\mbox{.}(2020)]%
        {DBLP:conf/eccv/MildenhallSTBRN20}
\bibfield{author}{\bibinfo{person}{Ben Mildenhall}, \bibinfo{person}{Pratul~P. Srinivasan}, \bibinfo{person}{Matthew Tancik}, \bibinfo{person}{Jonathan~T. Barron}, \bibinfo{person}{Ravi Ramamoorthi}, {and} \bibinfo{person}{Ren Ng}.} \bibinfo{year}{2020}\natexlab{}.
\newblock \showarticletitle{NeRF: Representing Scenes as Neural Radiance Fields for View Synthesis}. In \bibinfo{booktitle}{\emph{Computer Vision - {ECCV} 2020 - 16th European Conference, Glasgow, UK, August 23-28, 2020, Proceedings, Part {I}}} \emph{(\bibinfo{series}{Lecture Notes in Computer Science}, Vol.~\bibinfo{volume}{12346})}, \bibfield{editor}{\bibinfo{person}{Andrea Vedaldi}, \bibinfo{person}{Horst Bischof}, \bibinfo{person}{Thomas Brox}, {and} \bibinfo{person}{Jan{-}Michael Frahm}} (Eds.). \bibinfo{publisher}{Springer}, \bibinfo{pages}{405--421}.
\newblock
\urldef\tempurl%
\url{https://doi.org/10.1007/978-3-030-58452-8\_24}
\showDOI{\tempurl}


\bibitem[M{\"{u}}ller et~al\mbox{.}(2022)]%
        {DBLP:journals/tog/MullerESK22}
\bibfield{author}{\bibinfo{person}{Thomas M{\"{u}}ller}, \bibinfo{person}{Alex Evans}, \bibinfo{person}{Christoph Schied}, {and} \bibinfo{person}{Alexander Keller}.} \bibinfo{year}{2022}\natexlab{}.
\newblock \showarticletitle{Instant neural graphics primitives with a multiresolution hash encoding}.
\newblock \bibinfo{journal}{\emph{{ACM} Trans. Graph.}} \bibinfo{volume}{41}, \bibinfo{number}{4} (\bibinfo{year}{2022}), \bibinfo{pages}{102:1--102:15}.
\newblock
\urldef\tempurl%
\url{https://doi.org/10.1145/3528223.3530127}
\showDOI{\tempurl}


\bibitem[M{\"{u}}ller et~al\mbox{.}(2017)]%
        {DBLP:journals/cgf/MullerGN17}
\bibfield{author}{\bibinfo{person}{Thomas M{\"{u}}ller}, \bibinfo{person}{Markus~H. Gross}, {and} \bibinfo{person}{Jan Nov{\'{a}}k}.} \bibinfo{year}{2017}\natexlab{}.
\newblock \showarticletitle{Practical Path Guiding for Efficient Light-Transport Simulation}.
\newblock \bibinfo{journal}{\emph{Comput. Graph. Forum}} \bibinfo{volume}{36}, \bibinfo{number}{4} (\bibinfo{year}{2017}), \bibinfo{pages}{91--100}.
\newblock
\urldef\tempurl%
\url{https://doi.org/10.1111/CGF.13227}
\showDOI{\tempurl}


\bibitem[Nicolet et~al\mbox{.}(2023)]%
        {DBLP:journals/tog/NicoletRNKJM23}
\bibfield{author}{\bibinfo{person}{Baptiste Nicolet}, \bibinfo{person}{Fabrice Rousselle}, \bibinfo{person}{Jan Nov{\'{a}}k}, \bibinfo{person}{Alexander Keller}, \bibinfo{person}{Wenzel Jakob}, {and} \bibinfo{person}{Thomas M{\"{u}}ller}.} \bibinfo{year}{2023}\natexlab{}.
\newblock \showarticletitle{Recursive Control Variates for Inverse Rendering}.
\newblock \bibinfo{journal}{\emph{{ACM} Trans. Graph.}} \bibinfo{volume}{42}, \bibinfo{number}{4} (\bibinfo{year}{2023}), \bibinfo{pages}{62:1--62:13}.
\newblock
\urldef\tempurl%
\url{https://doi.org/10.1145/3592139}
\showDOI{\tempurl}


\bibitem[Nimier{-}David et~al\mbox{.}(2020)]%
        {DBLP:journals/tog/Nimier-DavidSRJ20}
\bibfield{author}{\bibinfo{person}{Merlin Nimier{-}David}, \bibinfo{person}{S{\'{e}}bastien Speierer}, \bibinfo{person}{Beno{\^{\i}}t Ruiz}, {and} \bibinfo{person}{Wenzel Jakob}.} \bibinfo{year}{2020}\natexlab{}.
\newblock \showarticletitle{Radiative backpropagation: an adjoint method for lightning-fast differentiable rendering}.
\newblock \bibinfo{journal}{\emph{{ACM} Trans. Graph.}} \bibinfo{volume}{39}, \bibinfo{number}{4} (\bibinfo{year}{2020}), \bibinfo{pages}{146}.
\newblock
\urldef\tempurl%
\url{https://doi.org/10.1145/3386569.3392406}
\showDOI{\tempurl}


\bibitem[Paszke et~al\mbox{.}(2019)]%
        {DBLP:conf/nips/PaszkeGMLBCKLGA19}
\bibfield{author}{\bibinfo{person}{Adam Paszke}, \bibinfo{person}{Sam Gross}, \bibinfo{person}{Francisco Massa}, \bibinfo{person}{Adam Lerer}, \bibinfo{person}{James Bradbury}, \bibinfo{person}{Gregory Chanan}, \bibinfo{person}{Trevor Killeen}, \bibinfo{person}{Zeming Lin}, \bibinfo{person}{Natalia Gimelshein}, \bibinfo{person}{Luca Antiga}, \bibinfo{person}{Alban Desmaison}, \bibinfo{person}{Andreas K{\"{o}}pf}, \bibinfo{person}{Edward~Z. Yang}, \bibinfo{person}{Zachary DeVito}, \bibinfo{person}{Martin Raison}, \bibinfo{person}{Alykhan Tejani}, \bibinfo{person}{Sasank Chilamkurthy}, \bibinfo{person}{Benoit Steiner}, \bibinfo{person}{Lu Fang}, \bibinfo{person}{Junjie Bai}, {and} \bibinfo{person}{Soumith Chintala}.} \bibinfo{year}{2019}\natexlab{}.
\newblock \showarticletitle{PyTorch: An Imperative Style, High-Performance Deep Learning Library}. In \bibinfo{booktitle}{\emph{Advances in Neural Information Processing Systems 32: Annual Conference on Neural Information Processing Systems 2019, NeurIPS 2019, December 8-14, 2019, Vancouver, BC, Canada}}, \bibfield{editor}{\bibinfo{person}{Hanna~M. Wallach}, \bibinfo{person}{Hugo Larochelle}, \bibinfo{person}{Alina Beygelzimer}, \bibinfo{person}{Florence d'Alch{\'{e}}{-}Buc}, \bibinfo{person}{Emily~B. Fox}, {and} \bibinfo{person}{Roman Garnett}} (Eds.). \bibinfo{pages}{8024--8035}.
\newblock
\urldef\tempurl%
\url{https://proceedings.neurips.cc/paper/2019/hash/bdbca288fee7f92f2bfa9f7012727740-Abstract.html}
\showURL{%
\tempurl}


\bibitem[Philip and Deschaintre(2023)]%
        {DBLP:journals/corr/abs-2305-02756}
\bibfield{author}{\bibinfo{person}{Julien Philip} {and} \bibinfo{person}{Valentin Deschaintre}.} \bibinfo{year}{2023}\natexlab{}.
\newblock \showarticletitle{Radiance Field Gradient Scaling for Unbiased Near-Camera Training}.
\newblock \bibinfo{journal}{\emph{CoRR}}  \bibinfo{volume}{abs/2305.02756} (\bibinfo{year}{2023}).
\newblock
\urldef\tempurl%
\url{https://doi.org/10.48550/ARXIV.2305.02756}
\showDOI{\tempurl}
\showeprint[arXiv]{2305.02756}


\bibitem[Qiao et~al\mbox{.}(2023)]%
        {DBLP:conf/iccv/QiaoGXF0L23}
\bibfield{author}{\bibinfo{person}{Yi{-}Ling Qiao}, \bibinfo{person}{Alexander Gao}, \bibinfo{person}{Yiran Xu}, \bibinfo{person}{Yue Feng}, \bibinfo{person}{Jia{-}Bin Huang}, {and} \bibinfo{person}{Ming~C. Lin}.} \bibinfo{year}{2023}\natexlab{}.
\newblock \showarticletitle{Dynamic Mesh-Aware Radiance Fields}. In \bibinfo{booktitle}{\emph{{IEEE/CVF} International Conference on Computer Vision, {ICCV} 2023, Paris, France, October 1-6, 2023}}. \bibinfo{publisher}{{IEEE}}, \bibinfo{pages}{385--396}.
\newblock
\urldef\tempurl%
\url{https://doi.org/10.1109/ICCV51070.2023.00042}
\showDOI{\tempurl}


\bibitem[Rosu and Behnke(2023)]%
        {DBLP:conf/cvpr/RosuB23}
\bibfield{author}{\bibinfo{person}{Radu~Alexandru Rosu} {and} \bibinfo{person}{Sven Behnke}.} \bibinfo{year}{2023}\natexlab{}.
\newblock \showarticletitle{PermutoSDF: Fast Multi-View Reconstruction with Implicit Surfaces Using Permutohedral Lattices}. In \bibinfo{booktitle}{\emph{{IEEE/CVF} Conference on Computer Vision and Pattern Recognition, {CVPR} 2023, Vancouver, BC, Canada, June 17-24, 2023}}. \bibinfo{publisher}{{IEEE}}, \bibinfo{pages}{8466--8475}.
\newblock
\urldef\tempurl%
\url{https://doi.org/10.1109/CVPR52729.2023.00818}
\showDOI{\tempurl}


\bibitem[Rudnev et~al\mbox{.}(2021)]%
        {DBLP:journals/corr/abs-2112-05140}
\bibfield{author}{\bibinfo{person}{Viktor Rudnev}, \bibinfo{person}{Mohamed Elgharib}, \bibinfo{person}{William A.~P. Smith}, \bibinfo{person}{Lingjie Liu}, \bibinfo{person}{Vladislav Golyanik}, {and} \bibinfo{person}{Christian Theobalt}.} \bibinfo{year}{2021}\natexlab{}.
\newblock \showarticletitle{Neural Radiance Fields for Outdoor Scene Relighting}.
\newblock \bibinfo{journal}{\emph{CoRR}}  \bibinfo{volume}{abs/2112.05140} (\bibinfo{year}{2021}).
\newblock
\showeprint[arXiv]{2112.05140}
\urldef\tempurl%
\url{https://arxiv.org/abs/2112.05140}
\showURL{%
\tempurl}


\bibitem[Srinivasan et~al\mbox{.}(2021)]%
        {DBLP:conf/cvpr/SrinivasanDZTMB21}
\bibfield{author}{\bibinfo{person}{Pratul~P. Srinivasan}, \bibinfo{person}{Boyang Deng}, \bibinfo{person}{Xiuming Zhang}, \bibinfo{person}{Matthew Tancik}, \bibinfo{person}{Ben Mildenhall}, {and} \bibinfo{person}{Jonathan~T. Barron}.} \bibinfo{year}{2021}\natexlab{}.
\newblock \showarticletitle{NeRV: Neural Reflectance and Visibility Fields for Relighting and View Synthesis}. In \bibinfo{booktitle}{\emph{{IEEE} Conference on Computer Vision and Pattern Recognition, {CVPR} 2021, virtual, June 19-25, 2021}}. \bibinfo{publisher}{Computer Vision Foundation / {IEEE}}, \bibinfo{pages}{7495--7504}.
\newblock
\urldef\tempurl%
\url{https://doi.org/10.1109/CVPR46437.2021.00741}
\showDOI{\tempurl}


\bibitem[Sun et~al\mbox{.}(2023)]%
        {DBLP:conf/iccv/0004CLYZM0Z023}
\bibfield{author}{\bibinfo{person}{Cheng Sun}, \bibinfo{person}{Guangyan Cai}, \bibinfo{person}{Zhengqin Li}, \bibinfo{person}{Kai Yan}, \bibinfo{person}{Cheng Zhang}, \bibinfo{person}{Carl Marshall}, \bibinfo{person}{Jia{-}Bin Huang}, \bibinfo{person}{Shuang Zhao}, {and} \bibinfo{person}{Zhao Dong}.} \bibinfo{year}{2023}\natexlab{}.
\newblock \showarticletitle{Neural-PBIR Reconstruction of Shape, Material, and Illumination}. In \bibinfo{booktitle}{\emph{{IEEE/CVF} International Conference on Computer Vision, {ICCV} 2023, Paris, France, October 1-6, 2023}}. \bibinfo{publisher}{{IEEE}}, \bibinfo{pages}{18000--18010}.
\newblock
\urldef\tempurl%
\url{https://doi.org/10.1109/ICCV51070.2023.01654}
\showDOI{\tempurl}


\bibitem[Tancik et~al\mbox{.}(2023)]%
        {DBLP:conf/siggraph/TancikWNLYWKASA23}
\bibfield{author}{\bibinfo{person}{Matthew Tancik}, \bibinfo{person}{Ethan Weber}, \bibinfo{person}{Evonne Ng}, \bibinfo{person}{Ruilong Li}, \bibinfo{person}{Brent Yi}, \bibinfo{person}{Terrance Wang}, \bibinfo{person}{Alexander Kristoffersen}, \bibinfo{person}{Jake Austin}, \bibinfo{person}{Kamyar Salahi}, \bibinfo{person}{Abhik Ahuja}, \bibinfo{person}{David McAllister}, \bibinfo{person}{Justin Kerr}, {and} \bibinfo{person}{Angjoo Kanazawa}.} \bibinfo{year}{2023}\natexlab{}.
\newblock \showarticletitle{Nerfstudio: {A} Modular Framework for Neural Radiance Field Development}. In \bibinfo{booktitle}{\emph{{ACM} {SIGGRAPH} 2023 Conference Proceedings, {SIGGRAPH} 2023, Los Angeles, CA, USA, August 6-10, 2023}}, \bibfield{editor}{\bibinfo{person}{Erik Brunvand}, \bibinfo{person}{Alla Sheffer}, {and} \bibinfo{person}{Michael Wimmer}} (Eds.). \bibinfo{publisher}{{ACM}}, \bibinfo{pages}{72:1--72:12}.
\newblock
\urldef\tempurl%
\url{https://doi.org/10.1145/3588432.3591516}
\showDOI{\tempurl}


\bibitem[Veach(1997)]%
        {DBLP:phd/us/Veach97}
\bibfield{author}{\bibinfo{person}{Eric Veach}.} \bibinfo{year}{1997}\natexlab{}.
\newblock \emph{\bibinfo{title}{Robust Monte Carlo methods for light transport simulation}}.
\newblock \bibinfo{thesistype}{Ph.\,D. Dissertation}. \bibinfo{school}{Stanford University, {USA}}.
\newblock
\urldef\tempurl%
\url{https://searchworks.stanford.edu/view/3911108}
\showURL{%
\tempurl}


\bibitem[Vicini et~al\mbox{.}(2021)]%
        {DBLP:journals/tog/ViciniSJ21}
\bibfield{author}{\bibinfo{person}{Delio Vicini}, \bibinfo{person}{S{\'{e}}bastien Speierer}, {and} \bibinfo{person}{Wenzel Jakob}.} \bibinfo{year}{2021}\natexlab{}.
\newblock \showarticletitle{Path replay backpropagation: differentiating light paths using constant memory and linear time}.
\newblock \bibinfo{journal}{\emph{{ACM} Trans. Graph.}} \bibinfo{volume}{40}, \bibinfo{number}{4} (\bibinfo{year}{2021}), \bibinfo{pages}{108:1--108:14}.
\newblock
\urldef\tempurl%
\url{https://doi.org/10.1145/3450626.3459804}
\showDOI{\tempurl}


\bibitem[Vicini et~al\mbox{.}(2022)]%
        {DBLP:journals/tog/ViciniSJ22}
\bibfield{author}{\bibinfo{person}{Delio Vicini}, \bibinfo{person}{S{\'{e}}bastien Speierer}, {and} \bibinfo{person}{Wenzel Jakob}.} \bibinfo{year}{2022}\natexlab{}.
\newblock \showarticletitle{Differentiable signed distance function rendering}.
\newblock \bibinfo{journal}{\emph{{ACM} Trans. Graph.}} \bibinfo{volume}{41}, \bibinfo{number}{4} (\bibinfo{year}{2022}), \bibinfo{pages}{125:1--125:18}.
\newblock
\urldef\tempurl%
\url{https://doi.org/10.1145/3528223.3530139}
\showDOI{\tempurl}


\bibitem[Wang et~al\mbox{.}(2021)]%
        {DBLP:conf/nips/WangLLTKW21}
\bibfield{author}{\bibinfo{person}{Peng Wang}, \bibinfo{person}{Lingjie Liu}, \bibinfo{person}{Yuan Liu}, \bibinfo{person}{Christian Theobalt}, \bibinfo{person}{Taku Komura}, {and} \bibinfo{person}{Wenping Wang}.} \bibinfo{year}{2021}\natexlab{}.
\newblock \showarticletitle{NeuS: Learning Neural Implicit Surfaces by Volume Rendering for Multi-view Reconstruction}. In \bibinfo{booktitle}{\emph{Advances in Neural Information Processing Systems 34: Annual Conference on Neural Information Processing Systems 2021, NeurIPS 2021, December 6-14, 2021, virtual}}, \bibfield{editor}{\bibinfo{person}{Marc'Aurelio Ranzato}, \bibinfo{person}{Alina Beygelzimer}, \bibinfo{person}{Yann~N. Dauphin}, \bibinfo{person}{Percy Liang}, {and} \bibinfo{person}{Jennifer~Wortman Vaughan}} (Eds.). \bibinfo{pages}{27171--27183}.
\newblock
\urldef\tempurl%
\url{https://proceedings.neurips.cc/paper/2021/hash/e41e164f7485ec4a28741a2d0ea41c74-Abstract.html}
\showURL{%
\tempurl}


\bibitem[Wang et~al\mbox{.}(2023b)]%
        {DBLP:journals/tog/WangWWZ23}
\bibfield{author}{\bibinfo{person}{Yu{-}Chen Wang}, \bibinfo{person}{Chris Wyman}, \bibinfo{person}{Lifan Wu}, {and} \bibinfo{person}{Shuang Zhao}.} \bibinfo{year}{2023}\natexlab{b}.
\newblock \showarticletitle{Amortizing Samples in Physics-Based Inverse Rendering Using ReSTIR}.
\newblock \bibinfo{journal}{\emph{{ACM} Trans. Graph.}} \bibinfo{volume}{42}, \bibinfo{number}{6} (\bibinfo{year}{2023}), \bibinfo{pages}{214:1--214:17}.
\newblock
\urldef\tempurl%
\url{https://doi.org/10.1145/3618331}
\showDOI{\tempurl}


\bibitem[Wang et~al\mbox{.}(2023a)]%
        {DBLP:journals/tog/WangSNSGKFMG23}
\bibfield{author}{\bibinfo{person}{Zian Wang}, \bibinfo{person}{Tianchang Shen}, \bibinfo{person}{Merlin Nimier{-}David}, \bibinfo{person}{Nicholas Sharp}, \bibinfo{person}{Jun Gao}, \bibinfo{person}{Alexander Keller}, \bibinfo{person}{Sanja Fidler}, \bibinfo{person}{Thomas M{\"{u}}ller}, {and} \bibinfo{person}{Zan Gojcic}.} \bibinfo{year}{2023}\natexlab{a}.
\newblock \showarticletitle{Adaptive Shells for Efficient Neural Radiance Field Rendering}.
\newblock \bibinfo{journal}{\emph{{ACM} Trans. Graph.}} \bibinfo{volume}{42}, \bibinfo{number}{6} (\bibinfo{year}{2023}), \bibinfo{pages}{260:1--260:15}.
\newblock
\urldef\tempurl%
\url{https://doi.org/10.1145/3618390}
\showDOI{\tempurl}


\bibitem[Wu et~al\mbox{.}(2023)]%
        {DBLP:conf/cvpr/WuHLZF023}
\bibfield{author}{\bibinfo{person}{Haoqian Wu}, \bibinfo{person}{Zhipeng Hu}, \bibinfo{person}{Lincheng Li}, \bibinfo{person}{Yongqiang Zhang}, \bibinfo{person}{Changjie Fan}, {and} \bibinfo{person}{Xin Yu}.} \bibinfo{year}{2023}\natexlab{}.
\newblock \showarticletitle{NeFII: Inverse Rendering for Reflectance Decomposition with Near-Field Indirect Illumination}. In \bibinfo{booktitle}{\emph{{IEEE/CVF} Conference on Computer Vision and Pattern Recognition, {CVPR} 2023, Vancouver, BC, Canada, June 17-24, 2023}}. \bibinfo{publisher}{{IEEE}}, \bibinfo{pages}{4295--4304}.
\newblock
\urldef\tempurl%
\url{https://doi.org/10.1109/CVPR52729.2023.00418}
\showDOI{\tempurl}


\bibitem[Xu et~al\mbox{.}(2023a)]%
        {DBLP:conf/siggrapha/XuALGB0BPKBLZR23}
\bibfield{author}{\bibinfo{person}{Linning Xu}, \bibinfo{person}{Vasu Agrawal}, \bibinfo{person}{William Laney}, \bibinfo{person}{Tony Garcia}, \bibinfo{person}{Aayush Bansal}, \bibinfo{person}{Changil Kim}, \bibinfo{person}{Samuel~Rota Bul{\`{o}}}, \bibinfo{person}{Lorenzo Porzi}, \bibinfo{person}{Peter Kontschieder}, \bibinfo{person}{Aljaz Bozic}, \bibinfo{person}{Dahua Lin}, \bibinfo{person}{Michael Zollh{\"{o}}fer}, {and} \bibinfo{person}{Christian Richardt}.} \bibinfo{year}{2023}\natexlab{a}.
\newblock \showarticletitle{VR-NeRF: High-Fidelity Virtualized Walkable Spaces}. In \bibinfo{booktitle}{\emph{{SIGGRAPH} Asia 2023 Conference Papers, {SA} 2023, Sydney, NSW, Australia, December 12-15, 2023}}, \bibfield{editor}{\bibinfo{person}{June Kim}, \bibinfo{person}{Ming~C. Lin}, {and} \bibinfo{person}{Bernd Bickel}} (Eds.). \bibinfo{publisher}{{ACM}}, \bibinfo{pages}{43:1--43:12}.
\newblock
\urldef\tempurl%
\url{https://doi.org/10.1145/3610548.3618139}
\showDOI{\tempurl}


\bibitem[Xu et~al\mbox{.}(2023b)]%
        {DBLP:journals/tog/XuBLZ23}
\bibfield{author}{\bibinfo{person}{Peiyu Xu}, \bibinfo{person}{Sai~Praveen Bangaru}, \bibinfo{person}{Tzu{-}Mao Li}, {and} \bibinfo{person}{Shuang Zhao}.} \bibinfo{year}{2023}\natexlab{b}.
\newblock \showarticletitle{Warped-Area Reparameterization of Differential Path Integrals}.
\newblock \bibinfo{journal}{\emph{{ACM} Trans. Graph.}} \bibinfo{volume}{42}, \bibinfo{number}{6} (\bibinfo{year}{2023}), \bibinfo{pages}{213:1--213:18}.
\newblock
\urldef\tempurl%
\url{https://doi.org/10.1145/3618330}
\showDOI{\tempurl}


\bibitem[Yang et~al\mbox{.}(2022a)]%
        {DBLP:conf/eccv/YangCCCW22}
\bibfield{author}{\bibinfo{person}{Wenqi Yang}, \bibinfo{person}{Guanying Chen}, \bibinfo{person}{Chaofeng Chen}, \bibinfo{person}{Zhenfang Chen}, {and} \bibinfo{person}{Kwan{-}Yee~K. Wong}.} \bibinfo{year}{2022}\natexlab{a}.
\newblock \showarticletitle{PS-NeRF: Neural Inverse Rendering for Multi-view Photometric Stereo}. In \bibinfo{booktitle}{\emph{Computer Vision - {ECCV} 2022 - 17th European Conference, Tel Aviv, Israel, October 23-27, 2022, Proceedings, Part {I}}} \emph{(\bibinfo{series}{Lecture Notes in Computer Science}, Vol.~\bibinfo{volume}{13661})}, \bibfield{editor}{\bibinfo{person}{Shai Avidan}, \bibinfo{person}{Gabriel~J. Brostow}, \bibinfo{person}{Moustapha Ciss{\'{e}}}, \bibinfo{person}{Giovanni~Maria Farinella}, {and} \bibinfo{person}{Tal Hassner}} (Eds.). \bibinfo{publisher}{Springer}, \bibinfo{pages}{266--284}.
\newblock
\urldef\tempurl%
\url{https://doi.org/10.1007/978-3-031-19769-7\_16}
\showDOI{\tempurl}


\bibitem[Yang et~al\mbox{.}(2022b)]%
        {DBLP:conf/nips/YangCCCW22}
\bibfield{author}{\bibinfo{person}{Wenqi Yang}, \bibinfo{person}{Guanying Chen}, \bibinfo{person}{Chaofeng Chen}, \bibinfo{person}{Zhenfang Chen}, {and} \bibinfo{person}{Kwan{-}Yee~K. Wong}.} \bibinfo{year}{2022}\natexlab{b}.
\newblock \showarticletitle{S\({}^{\mbox{3}}\)-NeRF: Neural Reflectance Field from Shading and Shadow under a Single Viewpoint}. In \bibinfo{booktitle}{\emph{Advances in Neural Information Processing Systems 35: Annual Conference on Neural Information Processing Systems 2022, NeurIPS 2022, New Orleans, LA, USA, November 28 - December 9, 2022}}, \bibfield{editor}{\bibinfo{person}{Sanmi Koyejo}, \bibinfo{person}{S.~Mohamed}, \bibinfo{person}{A.~Agarwal}, \bibinfo{person}{Danielle Belgrave}, \bibinfo{person}{K.~Cho}, {and} \bibinfo{person}{A.~Oh}} (Eds.).
\newblock
\urldef\tempurl%
\url{http://papers.nips.cc/paper\_files/paper/2022/hash/0a630402ee92620dc2de3b704181de9b-Abstract-Conference.html}
\showURL{%
\tempurl}


\bibitem[Yariv et~al\mbox{.}(2023)]%
        {DBLP:conf/siggraph/YarivHRVSSBM23}
\bibfield{author}{\bibinfo{person}{Lior Yariv}, \bibinfo{person}{Peter Hedman}, \bibinfo{person}{Christian Reiser}, \bibinfo{person}{Dor Verbin}, \bibinfo{person}{Pratul~P. Srinivasan}, \bibinfo{person}{Richard Szeliski}, \bibinfo{person}{Jonathan~T. Barron}, {and} \bibinfo{person}{Ben Mildenhall}.} \bibinfo{year}{2023}\natexlab{}.
\newblock \showarticletitle{BakedSDF: Meshing Neural SDFs for Real-Time View Synthesis}. In \bibinfo{booktitle}{\emph{{ACM} {SIGGRAPH} 2023 Conference Proceedings, {SIGGRAPH} 2023, Los Angeles, CA, USA, August 6-10, 2023}}, \bibfield{editor}{\bibinfo{person}{Erik Brunvand}, \bibinfo{person}{Alla Sheffer}, {and} \bibinfo{person}{Michael Wimmer}} (Eds.). \bibinfo{publisher}{{ACM}}, \bibinfo{pages}{46:1--46:9}.
\newblock
\urldef\tempurl%
\url{https://doi.org/10.1145/3588432.3591536}
\showDOI{\tempurl}


\bibitem[Zeltner et~al\mbox{.}(2021)]%
        {DBLP:journals/tog/ZeltnerSGJ21}
\bibfield{author}{\bibinfo{person}{Tizian Zeltner}, \bibinfo{person}{S{\'{e}}bastien Speierer}, \bibinfo{person}{Iliyan Georgiev}, {and} \bibinfo{person}{Wenzel Jakob}.} \bibinfo{year}{2021}\natexlab{}.
\newblock \showarticletitle{Monte Carlo estimators for differential light transport}.
\newblock \bibinfo{journal}{\emph{{ACM} Trans. Graph.}} \bibinfo{volume}{40}, \bibinfo{number}{4} (\bibinfo{year}{2021}), \bibinfo{pages}{78:1--78:16}.
\newblock
\urldef\tempurl%
\url{https://doi.org/10.1145/3450626.3459807}
\showDOI{\tempurl}


\bibitem[Zhang et~al\mbox{.}(2020)]%
        {DBLP:journals/tog/ZhangMYGZ20}
\bibfield{author}{\bibinfo{person}{Cheng Zhang}, \bibinfo{person}{Bailey Miller}, \bibinfo{person}{Kai Yan}, \bibinfo{person}{Ioannis Gkioulekas}, {and} \bibinfo{person}{Shuang Zhao}.} \bibinfo{year}{2020}\natexlab{}.
\newblock \showarticletitle{Path-space differentiable rendering}.
\newblock \bibinfo{journal}{\emph{{ACM} Trans. Graph.}} \bibinfo{volume}{39}, \bibinfo{number}{4} (\bibinfo{year}{2020}), \bibinfo{pages}{143}.
\newblock
\urldef\tempurl%
\url{https://doi.org/10.1145/3386569.3392383}
\showDOI{\tempurl}


\bibitem[Zhang et~al\mbox{.}(2019)]%
        {DBLP:journals/tog/ZhangWZGRZ19}
\bibfield{author}{\bibinfo{person}{Cheng Zhang}, \bibinfo{person}{Lifan Wu}, \bibinfo{person}{Changxi Zheng}, \bibinfo{person}{Ioannis Gkioulekas}, \bibinfo{person}{Ravi Ramamoorthi}, {and} \bibinfo{person}{Shuang Zhao}.} \bibinfo{year}{2019}\natexlab{}.
\newblock \showarticletitle{A differential theory of radiative transfer}.
\newblock \bibinfo{journal}{\emph{{ACM} Trans. Graph.}} \bibinfo{volume}{38}, \bibinfo{number}{6} (\bibinfo{year}{2019}), \bibinfo{pages}{227:1--227:16}.
\newblock
\urldef\tempurl%
\url{https://doi.org/10.1145/3355089.3356522}
\showDOI{\tempurl}


\bibitem[Zhang et~al\mbox{.}(2022a)]%
        {DBLP:conf/cvpr/ZhangLLS22}
\bibfield{author}{\bibinfo{person}{Kai Zhang}, \bibinfo{person}{Fujun Luan}, \bibinfo{person}{Zhengqi Li}, {and} \bibinfo{person}{Noah Snavely}.} \bibinfo{year}{2022}\natexlab{a}.
\newblock \showarticletitle{{IRON:} Inverse Rendering by Optimizing Neural SDFs and Materials from Photometric Images}. In \bibinfo{booktitle}{\emph{{IEEE/CVF} Conference on Computer Vision and Pattern Recognition, {CVPR} 2022, New Orleans, LA, USA, June 18-24, 2022}}. \bibinfo{publisher}{{IEEE}}, \bibinfo{pages}{5555--5564}.
\newblock
\urldef\tempurl%
\url{https://doi.org/10.1109/CVPR52688.2022.00548}
\showDOI{\tempurl}


\bibitem[Zhang et~al\mbox{.}(2021)]%
        {DBLP:conf/cvpr/ZhangLWBS21}
\bibfield{author}{\bibinfo{person}{Kai Zhang}, \bibinfo{person}{Fujun Luan}, \bibinfo{person}{Qianqian Wang}, \bibinfo{person}{Kavita Bala}, {and} \bibinfo{person}{Noah Snavely}.} \bibinfo{year}{2021}\natexlab{}.
\newblock \showarticletitle{PhySG: Inverse Rendering With Spherical Gaussians for Physics-Based Material Editing and Relighting}. In \bibinfo{booktitle}{\emph{{IEEE} Conference on Computer Vision and Pattern Recognition, {CVPR} 2021, virtual, June 19-25, 2021}}. \bibinfo{publisher}{Computer Vision Foundation / {IEEE}}, \bibinfo{pages}{5453--5462}.
\newblock
\urldef\tempurl%
\url{https://doi.org/10.1109/CVPR46437.2021.00541}
\showDOI{\tempurl}


\bibitem[Zhang et~al\mbox{.}(2022b)]%
        {DBLP:conf/cvpr/ZhangSHFJZ22}
\bibfield{author}{\bibinfo{person}{Yuanqing Zhang}, \bibinfo{person}{Jiaming Sun}, \bibinfo{person}{Xingyi He}, \bibinfo{person}{Huan Fu}, \bibinfo{person}{Rongfei Jia}, {and} \bibinfo{person}{Xiaowei Zhou}.} \bibinfo{year}{2022}\natexlab{b}.
\newblock \showarticletitle{Modeling Indirect Illumination for Inverse Rendering}. In \bibinfo{booktitle}{\emph{{IEEE/CVF} Conference on Computer Vision and Pattern Recognition, {CVPR} 2022, New Orleans, LA, USA, June 18-24, 2022}}. \bibinfo{publisher}{{IEEE}}, \bibinfo{pages}{18622--18631}.
\newblock
\urldef\tempurl%
\url{https://doi.org/10.1109/CVPR52688.2022.01809}
\showDOI{\tempurl}


\bibitem[Zhu et~al\mbox{.}(2021)]%
        {DBLP:journals/tog/ZhuBXBV0SHY21}
\bibfield{author}{\bibinfo{person}{Junqiu Zhu}, \bibinfo{person}{Yaoyi Bai}, \bibinfo{person}{Zilin Xu}, \bibinfo{person}{Steve Bako}, \bibinfo{person}{Edgar Vel{\'{a}}zquez{-}Armend{\'{a}}riz}, \bibinfo{person}{Lu Wang}, \bibinfo{person}{Pradeep Sen}, \bibinfo{person}{Milos Hasan}, {and} \bibinfo{person}{Ling{-}Qi Yan}.} \bibinfo{year}{2021}\natexlab{}.
\newblock \showarticletitle{Neural complex luminaires: representation and rendering}.
\newblock \bibinfo{journal}{\emph{{ACM} Trans. Graph.}} \bibinfo{volume}{40}, \bibinfo{number}{4} (\bibinfo{year}{2021}), \bibinfo{pages}{57:1--57:12}.
\newblock
\urldef\tempurl%
\url{https://doi.org/10.1145/3450626.3459798}
\showDOI{\tempurl}


\end{thebibliography}

\end{document}